\documentclass[11pt,a4paper]{article}
\usepackage[hyperref]{acl2020}
\usepackage{times}
\usepackage{latexsym}
\usepackage{enumitem}

\usepackage{graphicx}
\usepackage{dirtytalk}
\usepackage{multirow}
\usepackage{adjustbox}
\usepackage{flushend}

\usepackage{microtype}

\title{ImageArg: A Multi-modal Tweet Dataset for Image Persuasiveness Mining}

\author{Anonymous Argument Mining Workshop submission}
\author{Zhexiong Liu$^*$, Meiqi Guo$^*$, Yue Dai$^*$, Diane Litman \\
Department of Computer Science \\ University of Pittsburgh, 
Pittsburgh, Pennsylvania, 15260 \\
\texttt{\{zhexiong.liu,meiqi.guo,yud42,dlitman\}@pitt.edu}}

\date{}

\begin{document}
\maketitle
\begin{abstract}
% Our project is a corpus annotation project. While many annotation studies have been conducted on text annotating for argument mining tasks, no previous works explored the roles of images within documents in the persuasiveness perspective. Being aware of the growing amount of images contained within social media, and the importance of these images in aspect of argumentation, we propose a corpus annotation project that labels images within online documents and annotates the content as well as the persuasion mode of images. We evaluate the quality and effectiveness of our annotation scheme. With the collected corpus, we will conduct an insight analysis as well as building a multi-modality argumentation mining system.

The growing interest in developing  corpora of persuasive texts has promoted  applications in automated systems, e.g., debating and essay scoring systems; however, there is little prior work mining image persuasiveness from an argumentative perspective. To expand persuasiveness mining into a multi-modal realm, we present a multi-modal dataset, \textit{ImageArg}, consisting of annotations of image persuasiveness in tweets. The annotations are based on a persuasion taxonomy we developed to  explore image functionalities and the means of persuasion. We benchmark image persuasiveness tasks on \textit{ImageArg} using widely-used multi-modal learning methods. The experimental results show that our dataset offers a useful resource for this rich and challenging topic, and there is ample room for modeling improvement.  
\end{abstract}

\def\thefootnote{*}\footnotetext{These authors contributed equally to this work.}
\def\thefootnote{\arabic{footnote}}

\section{Introduction}
Argumentation mining (AM) aims to analyze authors' argumentative stance by automatically identifying argumentative structures and their relationships \cite{green2014proceedings}. 
As a fundamental component in AM, computational persuasiveness analysis has gained considerable momentum  due to growing resources and downstream applications \cite{chatterjee2006word,park2014computational,wei2016post,lukin2017argument,chakrabarty2017context,lytos2019evolution}. 
Aiming at automatically evaluating how well one party can change another party's opinions or behaviors, computational persuasiveness tasks are critical yet challenging. 

\begin{figure}[t]
    \centering
    \includegraphics[width=1\columnwidth]{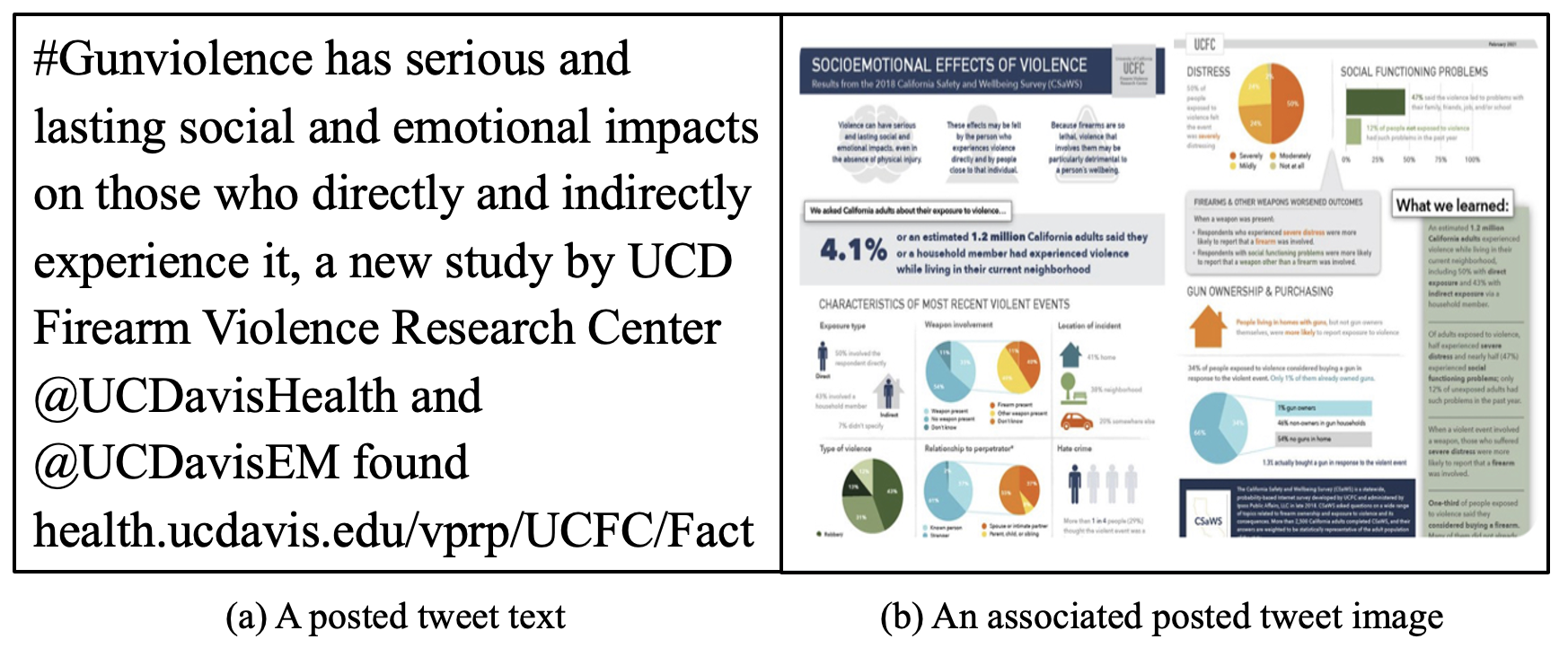}
    \caption{(a) The tweet text uses gun violence to argue for \textit{gun control}. (b) The image makes the argument more persuasive by providing supplementary statistics relating violence to gun ownership in California.}
    \label{fig:example}
\end{figure}

Recent work in AM has brought attention to mining persuasiveness in essays. \citet{stab2014annotating} and \citet{habernal2017argumentation} developed the Argument Annotated Essays Corpus (AAEC) where stance, argument components, and argumentative relations were annotated. \citet{carlile2018give} extended AAEC annotations with persuasiveness scores, as well as with argumentative attributes that potentially impact persuasiveness (Eloquence, Specificity, Relevance, and Evidence) and the means of persuasion (Ethos, Pathos, or Logos). These are all text-based annotations, however,  missing the opportunity to leverage other modalities (e.g., images) that potentially enhance the persuasiveness of the argument. For example, the image showing statistic charts in Fig. \ref{fig:example} makes the tweet text more convincing. %Nevertheless, image persuasiveness has rarely been explored in the AM community.  
To address the gap that image persuasivness has rarely been explored in the AM community, we create a new multi-modal dataset, \textit{ImageArg}, that annotates image persuasiveness in tweets and extends persuasiveness mining to a multi-modal realm. 
% Nevertheless, many online documents on social media heavily rely on accompanying images to express their arguments. Moreover, these non-text contents potentially play an essential role in the argumentative structures. In some cases, these non-text contents might enhance the arguments within the texts (e.g., supporting texts with statistical charts
% % , as shown in Figure \ref{fig:example}
% ), and in some cases, they could ultimately reverse the stances and opinions represented by the texts (e.g., ironic drawing that delivers arguments utterly opposite to the text). 
% Although images express rich visual stances, they receive less attention and are rarely explored in computational persuasiveness tasks.
% Although images express rich visual stances but received less attention. For example, images are favored in social media to express visual stance towards an argumentation. 
% Figure \ref{fig:example} shows a tweet that utilizes chart in images to support gun control

Regarding \textit{ImageArg} construction, we first extend annotation schemes that are previously developed to capture the persuasive strength of text arguments in AAEC \cite{duthie2016mining, wachsmuth2018argumentation, carlile2018give} to a new modality of image. Specifically, we develop a novel strategy (Sec. \ref{Sec: Score}) to annotate multi-modal persuasiveness gains that measure if the persuasivness of a tweet's text increases after adding a visual image. Second, we devise a taxonomy to annotate image content (Sec. \ref{Sec: Content Type}) that explicitly identifies image functionalities from a persuasive perspective. Furthermore, we adapt existing text attributes %(Sec. \ref{Sec: Persuasion Mode}) 
used in \citet{carlile2018give} to annotate image persuasion modes (Sec. \ref{Sec: Persuasion Mode}) by exploring different annotation strategies (Sec. \ref{Sec: annotation_strategies}). 
We evaluate the inter-rater agreement on our proposed annotation schemes as well as the quality of the annotated samples.   

%Our contributions include: (1) We propose a novel strategy to annotate image persuasiveness and develop image-oriented schemes to identify image functionality and persuasion mode. (2) We create a new multi-modal dataset \textit{ImageArg} that contains annotations for image persuasiveness in tweets.
With \textit{ImageArg}, we first report the basic statistics of the dataset and conduct a thorough analysis between different annotation dimensions (Sec. \ref{Sec: statistics}). We observe a strong correlation between human political ideology (i.e. stance towards a social topic) and the argumentative features in their posted tweets, as well as mutual influences between image content and persuasion mode. In addition, we benchmark model performance on multiple argumentative classification tasks annotated in \textit{ImageArg} (Sec. \ref{Sec: quan_experiments}). Specifically, we employ multi-modal learning methods to classify stance, image persuasiveness, image content, and image persuasion mode. 
Our benchmark results highlight the challenge of these tasks and indicate there is ample room for model improvement. We demonstrate the limitation of these general multi-modal methods and discuss possible future work. We further conduct a qualitative study on a real-world application, retrieving the most persuasive images given a tweet text, by using our trained classifiers (Sec. \ref{Sec: qual_experiments}), which offers a starting point for developing an intelligent tool that recommends persuasive images to users based on their textual inputs. Our code and data is publicly available at: \url{https://github.com/MeiqiGuo/ArgMining2022-ImageArg}.

\section{Related Work}
\textbf{Computational Persuasiveness} 
While classical AM focuses on identifying argumentative components and their relations %presented in the natural language 
\cite{stab2014argumentation,stab2018cross, lawrence2020argument}, recent work has developed interest in persuasiveness related tasks \cite{chatterjee2014verbal, park2014computational, lukin2017argument, carlile2018give, chakrabarty2019ampersand}.
%For instance, 
In addition, \citet{riley1954communication}, \citet{o2015persuasion}, and \citet{wei2016post} investigate
%the persuasiveness 
ranking 
%ranks
debate arguments on the same topic based on their persuasiveness, but they failed to investigate the factors that make arguments persuasive. \citet{lukin2017argument} and \citet{persing2017can} examine how audience variables (e.g., personality) influence persuasiveness through different argument styles (e.g., factual vs. emotional arguments), but only focus on the text modality. %and has not considered image modality (e.g., image content).
% citet{hidey2017analyzing} investigate different semantic types regarding claims and premises.
\citet{higgins2012ethos} and \citet{carlile2018give} study the persuasion strategies, i.e., Ethos (credibility), Logos (reason), and Pathos (emotion), in the scope of reports or student essays.
%while \citet{al2016news} annotate a news editorial corpus with fine-grained argumentative discourse units for analyzing argumentation strategies. 
We follow their work developed for text corpora and extend the annotation schemes to the image modality. Although \citet{park2014computational}, \citet{joo2014visual}, and \citet{Huang2016InferringVP} utilize facial expressions and bodily gestures to analyze persuasiveness in social multimedia, their work is limited to the human portrait and fails to generalize to diverse image domains. Some prior work study persuasive advertisements in a multi-modal way \cite{Hussain_2017_Automatic, guo2021detecting}. Different from our argumentative mining goal, they focus on the sentiment, intent reasoning and persuasive strategies that are narrowly designed for ads. Thus, annotating a multi-modal tweet dataset focusing on image persuasiveness is under-explored in existing work, and has ample value for social science.  

\begin{figure*}[t]
    \centering
    \includegraphics[width=1\textwidth]{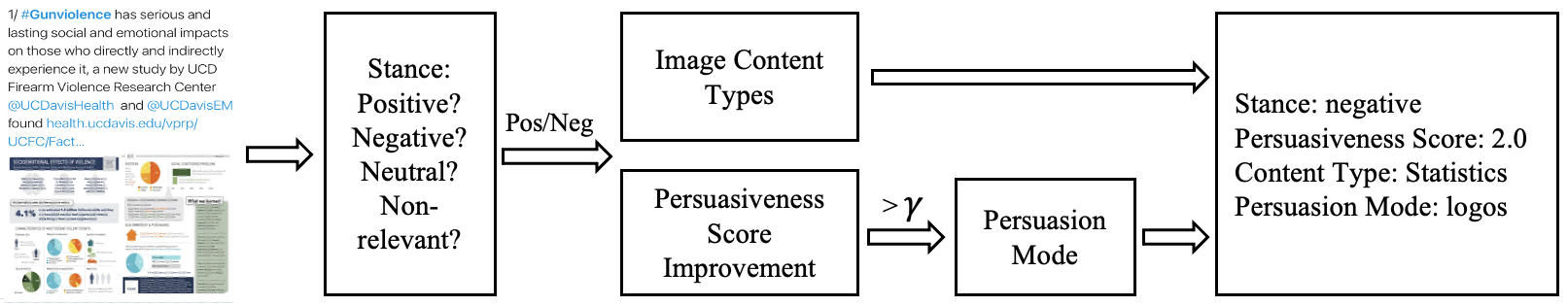}
    \caption{The overview of our annotation pipeline. Annotators start by annotating the argumentative stance of input tweets. Afterwards, tweets with either positive or negative stances are annotated for image content types and persuasiveness score improvement. The persuasion mode is further annotated if persuasiveness score improvement exceeds a given threshold $\gamma$. We use $\gamma=0.5$ when we annotate data and test with different $\gamma$ values for persuasiveness classification task (Table \ref{tab:persuasion_mode_threshold}).}
    \label{fig:pipeline} 
\end{figure*}

\textbf{Multi-modal Learning} The ability to process and understand multi-modal input for AI models has recently received much attention since the multi-modal signals are generally complementary for real-world applications \cite{aytar2016soundnet, Zhang_2018_Equal, alwassel2020self}. In the area of vision-language, tasks are mainly designed for evaluating models' ability to understand visual information as well as expressing the reasoning in language \cite{antol2015vqa, goyal2017making, hudson2019gqa}. In addition to the main stream, a few works study the relationship between image and text: \citet{alikhani2019cite} annotates the discourse relations between text and accompanying imagery in recipe instructions; and \citet{kruk2019integrating} investigates the multi-modal document intent in instagram posts. However, multi-modal learning for AM has been under-explored due to a lack of multi-modal corpora. This drives us to build {\it ImageArg} and to analyze the effectiveness of multi-modal learning on AM tasks. With respect to modeling, researchers focus on learning good representation of each modality and developing effective fusion methods  \cite{tsai2018learning, hu2019dense, tan2019lxmert, lu202012}. In this work, we establish a benchmark performance for {\it ImageArg} by using fundamental and common encoders and fusion methods.

%Recent studies have shown that multi-modal machine learning approaches can capture joint representations from multi-modalities and lead to better prediction performance \cite{atrey2010multimodal,baltruvsaitis2018multimodal}. Researchers have thus leveraged multimodal data to learn representations for several vision and language downstream tasks \cite{antol2015vqa, aytar2016soundnet, yang2017deep, faghri2017vse++, tsai2018learning, hu2019dense, alikhani2019cite, hudson2019gqa, kruk2019integrating, lu202012, alwassel2020self}. In addition to multimodal fusion, \citet{silberer2014learning} and \citet{gu2018look} learn multimodal embedding, while \citet{cao2016deep} introduce a cross-modality hashing approach for text and image retrieval in multi-modal cases. 

\section{Annotation Scheme}
\label{Corpus}

We propose an annotation scheme to capture an image's impact on the persuasiveness of multi-modal tweets. We build a corpus of Twitter posts on a social topic (e.g., \textit{gun control}), then annotate the image within each post along four dimensions. The annotation pipeline is shown in Fig. \ref{fig:pipeline}. First, we determine \textbf{(1)} the {\bf stance} of the entire tweet (Sec. \ref{Sec: Stance}). Specifically, we assume one tweet holds a consistent stance in its text and image since the author would intend to deliver a consistent argument. For those tweets annotated with a positive or negative stance, we also annotate \textbf{(2)} the {\bf persuasiveness scores} of the tweet image (Sec. \ref{Sec: Score}) and \textbf{(3)} the image {\bf content type}. % if it holds a positive or negative stance on the topic.
The content types identify image roles from an argumentative perspective (Sec. \ref{Sec: Content Type}). Finally, we \textbf{(4)} identify the {\bf persuasion mode} of an image that is annotated as persuasive. The persuasion mode indicates how the images persuade audiences (Sec. \ref{Sec: Persuasion Mode}). 
Note that with this annotation pipeline, all tweets will first be annotated for  stance. Then, only  tweets with a clear stance will be annotated for content type and persuasiveness scores. Finally, only  tweets where the images are persuasive will be annotated for  persuasion mode.
% We present our data acquisition and corpus creation in \ref{Sec: Corpus Creation}.

\begin{figure}[t]
    \centering
    \includegraphics[width=0.95\columnwidth]{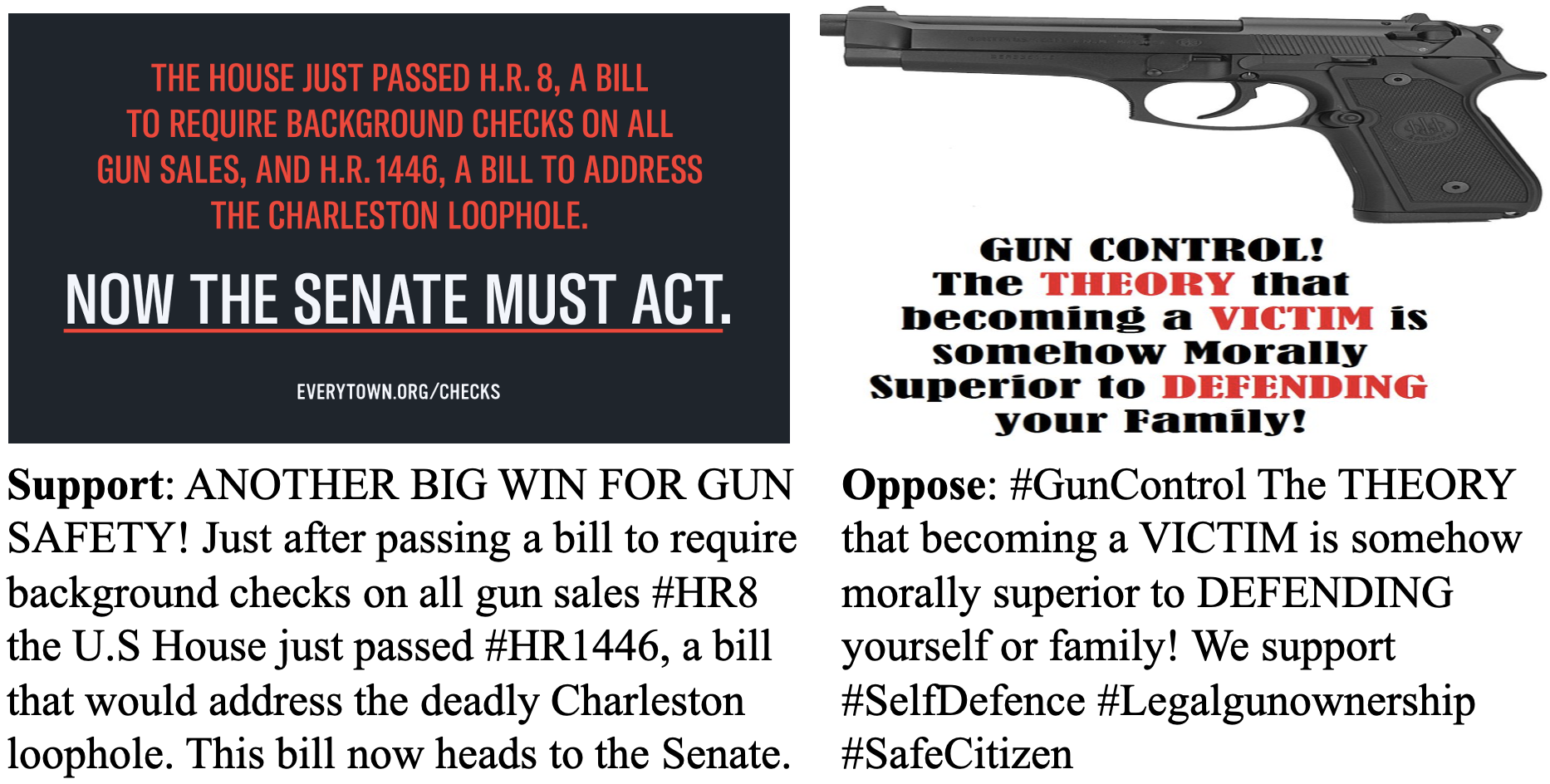}
    \caption{Examples of positive (support) and negative (oppose) tweets.}
    \label{fig:image-stance}
\end{figure}

\subsection{Stance}\label{Sec: Stance}
We use existing methods \cite{mohammad2017stance} to verify if the image holds a clear stance on a given topic.
% We filter out tweets irrelevant to the topics and conduct . In addition, since the stances contribute to extensive argumentation mining pipelines and potentially play an essential role in the persuasiveness-related tasks \cite{lytos2019evolution}, we label stances to investigate their potential impacts on the other attributes.
Specifically, given a tweet (including text and images), we ask annotators to select among four stances that are extended from \citet{mohammad2017stance}: positive (i.e., support), negative (i.e., oppose), neutral, or irrelevant to the topic. We continue with the next annotation steps only if a tweet holds a positive or negative stance. Otherwise, it is discarded for our persuasion study. %as an irrelevant post. 
We show examples in Fig. \ref{fig:image-stance}. 
% Our detailed coding manual for stance annotation is shown in Appendix \ref{Sec: stance code manual}.

\subsection{Image Persuasiveness Scores}\label{Sec: Score}
For a tweet that holds a positive or negative stance, we study the impact of its image by computing an image persuasiveness score improvment. We adopt five levels of text persuasiveness scores proposed in \citet{carlile2018give} in the annotation process: (L0) no persuasiveness (score = 0): the annotated target fails to convince the audience at all. (L1) medium persuasiveness (score = 1): the annotated target partially convinces the audience. (L2) persuasive (score = 2): the annotated target is convincing to the audience. (L3) high persuasiveness (score = 3): the annotated target is very convincing to the audience. (L4) extreme persuasiveness (score = 4): the annotated target is compelling to the audience.

% \begin{itemize}
% \setlength\itemsep{0.05em}
%     \item \textbf{Not persuasive at all (score = 0).} The annotated target fails to convince the audience at all.
%     \item \textbf{Somewhat persuasive (score = 1).} The annotated target partially convinces the audience.
%     \item \textbf{Persuasive (score = 2).} The annotated target is convincing to the audience.
%     \item \textbf{Quite persuasive (score = 3).} The annotated is very convincing to the audience.
%     \item \textbf{Extremely persuasive (score = 4).} The annotated target is compelling to the audience.
% \end{itemize}
Different from \citet{carlile2018give} that annotates the persuasiveness score directly, we propose a novel method to compute the image persuasiveness score. In particular, we calculate the differences with/without images to quantify image persuasiveness scores. We first ask annotators to choose one of 5 persuasiveness levels based on pure text from the tweet. Next, we ask annotators to give a second choice based on both text and image from the tweet.
Suppose each sample has three annotations and each annotation has two persuasiveness scores: one for the text-only ($s_t$), the other for the image-text ($s_{it}$). We compute persuasiveness score difference $\Delta s_i = max(s_{it} - s_t, 0)$ for each annotation, as the persuasiveness gain from the image. Then, we compute the average of the three annotations ($\Delta s_i$) as the final image persuasiveness score. To interpret image persuasiveness, we use a threshold ($\gamma$) that encodes the score into a binary label (i.e., persuasiveness or not). 
%Based on our pilot study, we select $\gamma=0.5$ for dataset annotation. 
If $\Delta s_i$ is higher than the threshold ($\gamma$), it indicates that adding an image improves tweet persuasiveness, thus the image is considered as persuasive.
%; otherwise, they are labeled  as non-persuasive. 
We show examples with different image persuasiveness scores in Fig. \ref{fig:image-score}. 
% We provide the detailed coding manual in Appendix \ref{Sec: content code manual}.

\subsection{Image Content Types}\label{Sec: Content Type}
% We defined six categories for representing the content of images: Statistics, Testimony, Anecdote, Slogan, Scene Photo and Symbolic Photo. 
For persuasive samples, we investigate their image argumentative roles. In particular, we annotate the image content types from an argumentative perspective to describe what kind of evidence images provide to improve  tweet persuasiveness (e.g., supportive data, authorized photos, etc.).
We leverage \citet{al2016news}'s definition of argumentative roles of evidence to categorize image content: Statistics, Testimony, and Anecdote. However, we notice that the categories fail to capture all the image contents that frequently appear in tweet posts, for example, photographs. To this end, we propose a Slogan category highlighting text in images, and also propose Scene photo and Symbolic photo categories regarding image content in the visual modality. More details are specified as follows: 
% with the coding manual provided in Appendix \ref{Sec: content code manual}: % ; Fig \ref{fig:image-role} shows some examples.
\begin{figure}[t]
    \centering
    \includegraphics[width=1\columnwidth]{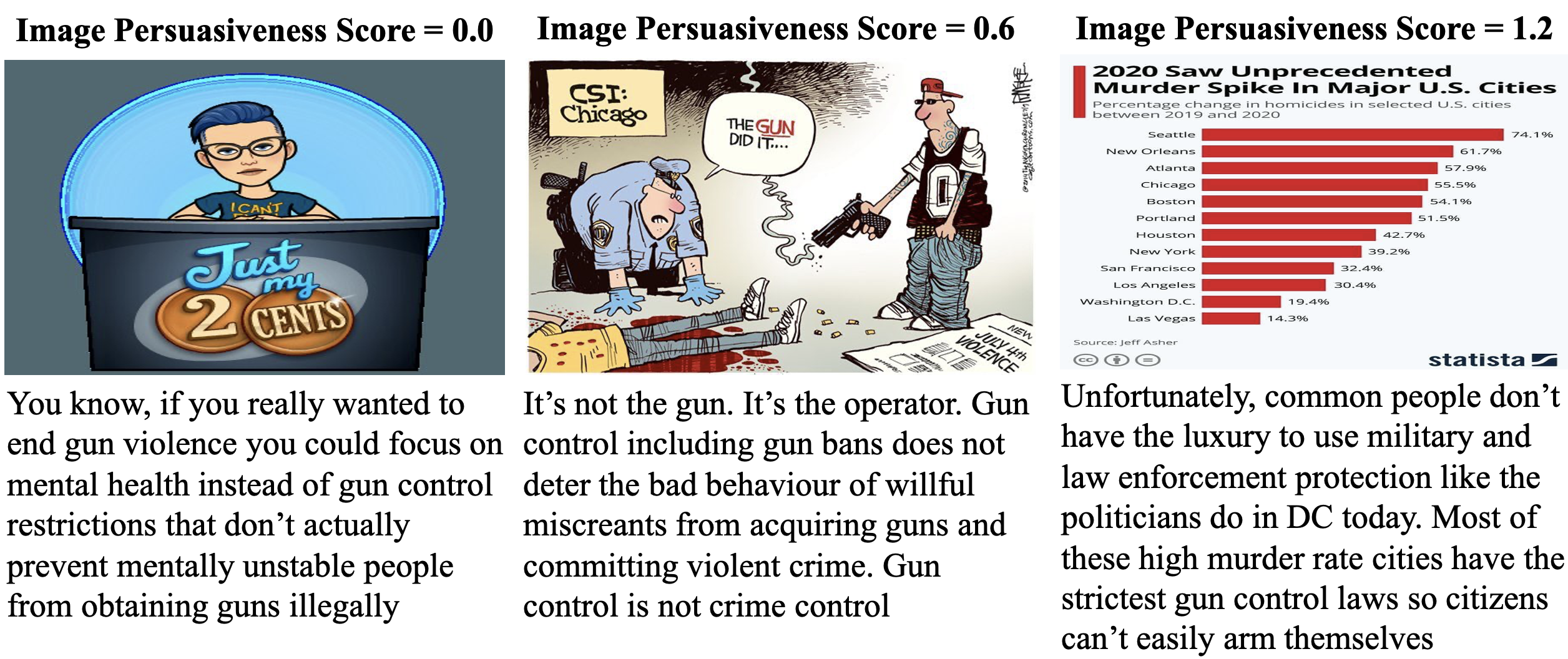}
    \caption{Examples of tweets with 0, 0.6, and 1.2 image persuasiveness scores.}
    \label{fig:image-score}
\end{figure}

\begin{figure}[t]
    \centering
    \includegraphics[width=1\columnwidth]{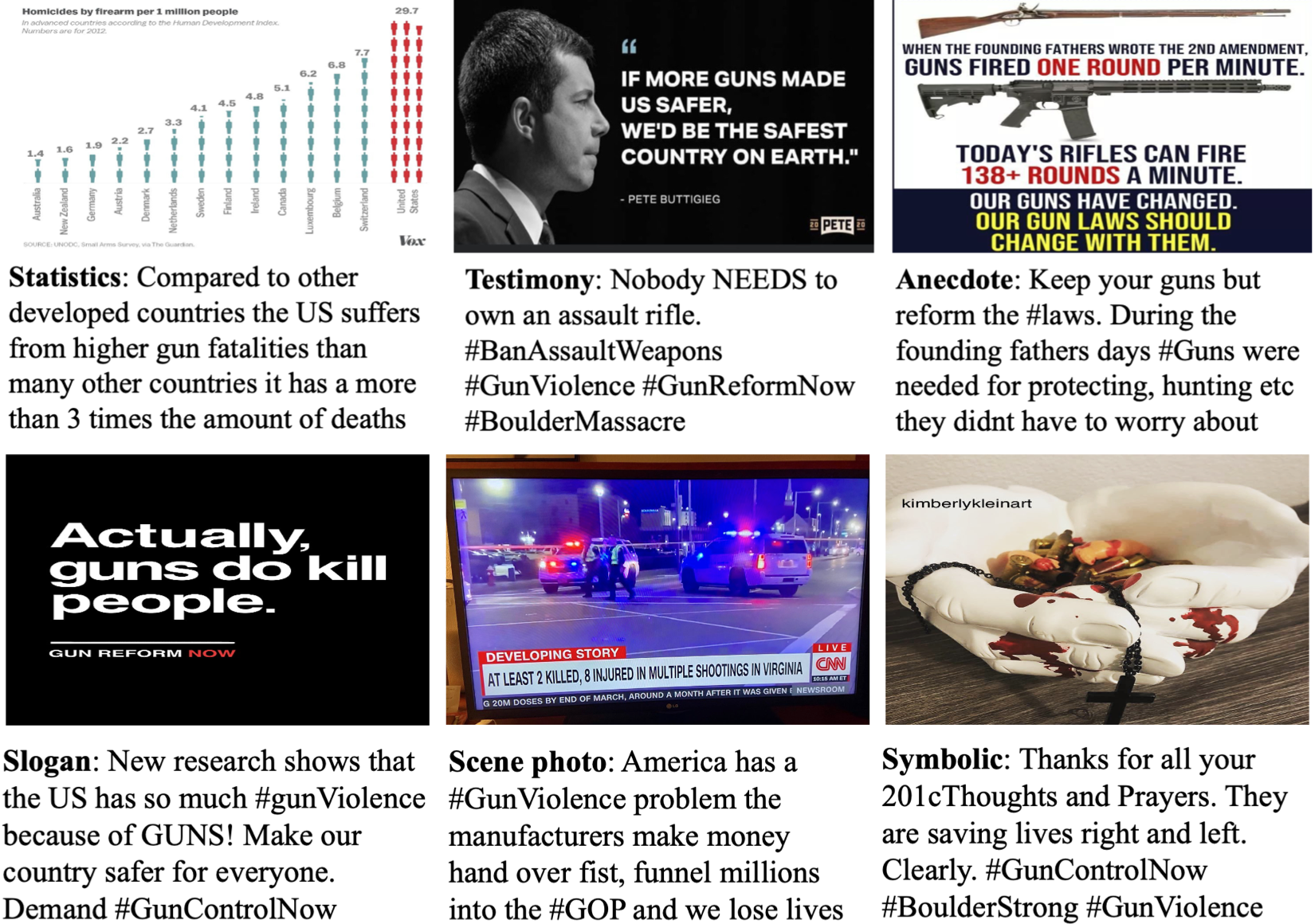}
    \caption{Examples of image content types in tweets: statistics, testimony, anecdote, slogan, scene photo, and symbolic.}
    \label{fig:image-role}
\end{figure}
\begin{figure}[t]
    \centering
    \includegraphics[width=1\columnwidth]{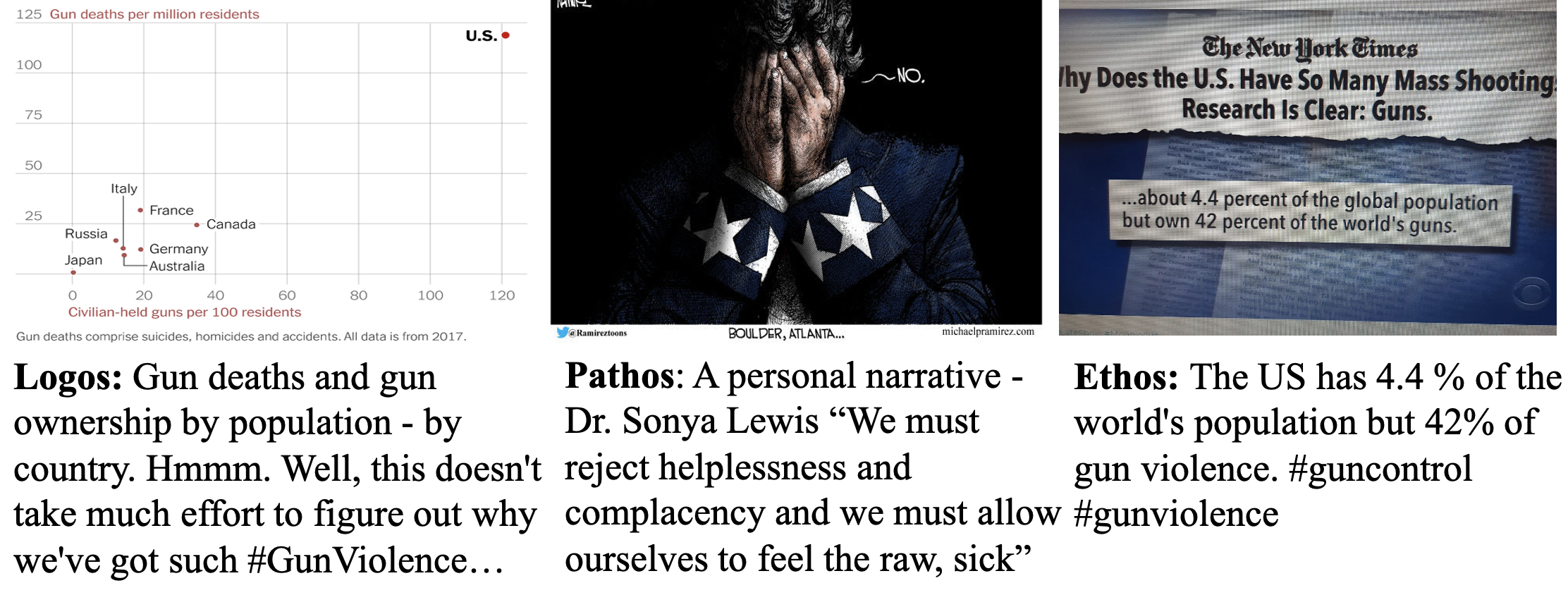}
    \caption{Examples of persuasion mode in tweet: logos, pathos, and ethos.}
    \label{fig:image-mode}
\end{figure}
\begin{itemize}[leftmargin=*]
\setlength\itemsep{-0.15em}
  \item
   \textbf{Statistics}: Images provide evidence by stating or quoting quantitative information, such as a chart or diagram showing data, that is related to the tweet text. 
  In Fig. \ref{fig:image-role}, the  image provides quantitative statistics on gun fatalities.
    \item 
    \textbf{Testimony}: Images quote statements or conclusions from an authority, such as a piece of articles or claims from an official document, that is related to the tweet text. 
    For example, in Fig. \ref{fig:image-role}, the testimony image cites a statement given by the transportation secretary.
    \item 
    \textbf{Anecdote}: Images provide information based on the author's personal experience, such as facts/personal stories, that are related to the tweet text. 
    In Fig. \ref{fig:image-role}, the anecdote image shows the fact that guns are developed since the period of the 2nd amendment, and therefore the laws for guns should be developed as well.
    \item 
    \textbf{Slogan}: Images embed pieces of advertising/slogan text. 
    In Fig. \ref{fig:image-role}, the slogan image presents a phrase \say{Actually guns do kill people. Gun Reform Now}.
    \item 
    \textbf{Scene photo}: Images show a real scene or photograph that is related to the tweet text. 
    In  Fig. \ref{fig:image-role}, the  image shows a photo of a gun violence scene reported by CNN news.
    \item 
    \textbf{Symbolic photo}: Images show a symbol/art that expresses the author's viewpoints in a non-literal way. In Fig. \ref{fig:image-role}, the symbolic photo shows a pair of artificial bloodied hands holding bullets and a cross which symbolically reveals the brutality of gun violence.
    
\end{itemize} 

\subsection{Image Persuasion Modes} \label{Sec: Persuasion Mode}
To investigate how images convince an audience (e.g., by providing strong logic, touching audiences emotionally, etc.),
we  annotate the persuasion modes of images by
%To prevent ambiguous definitions, we 
leveraging the %standard persuasion modes 
definitions in \citet{braet1992ethos} for Logos, Pathos, and Ethos. The modes form the rhetorical triangle, and both the textual and visual modalities follow these dimensions in the persuasiveness perspective. Fig. \ref{fig:image-mode} shows examples, details are specified below:
% while details are specified below and  the coding manual is provided in Appendix \ref{Sec: mode code manual}:
\begin{itemize}[leftmargin=*] 
\setlength\itemsep{-0.25em}
    \item 
    \textbf{Logos}: The image appeals to logic and reasoning, which persuades audiences with reasoning from a fact/statistics/study case/scientific evidence. In Fig. \ref{fig:image-mode}, the Logos image provides a chart that shows the high gun deaths and the high gun ownership by the population of the US, which implies a logical relationship between gun death and gun ownership.
    \item 
    \textbf{Pathos}: The image appeals to emotion, i.e., evokes emotional impact that leads to higher persuasiveness. In Fig. \ref{fig:image-mode}, the Pathos image provides art that shows the grieved ``Uncle Sam" saying ``no" with helplessness, which evokes the desire to \textit{gun control}.
    \item 
    \textbf{Ethos}: The image appeals to ethics, which enhances credibility and trustworthiness. In Fig. \ref{fig:image-mode}, the Ethos image takes a screenshot of the source of a report from New York Times, which increases credibility.
    
\end{itemize}

% \begin{figure}[t]
%     \centering
%     \includegraphics[scale=0.22]{figures/example1.png}
%     \caption{Tweet examples with image persuasive mode.}
%     \label{fig:image-mode}
% \end{figure}

% \section{Corpus Annotation}

\section{Corpus Creation}
\label{Sec: Corpus Creation}
\subsection{Data Collection}
\label{Sec: Gun Control Annotation}
% We collect our date from Twitter platform as it has rich image-accompanied posts across diverse topics. 
We collect raw tweets containing both image and text across 3 topics (\textit{gun control, immigration} and \textit{abortion}) used in \citet{mochales2011argumentation} and \citet{stab2018cross}. 
% For the course project, we evaluate our annotation scheme on 87 posts about the topic of \emph{gun control} which are extracted by the below pipeline.
Specifically, we retrieve tweets with images that contain pre-defined keywords\footnote{We use keywords provided in \citet{guo2020inflating}'s work.} through TwitterAPI\footnote{https://developer.twitter.com/en/docs/twitter-api}. The raw data (286k tweets) are collected in a two-year window from 3/29/2019 to 3/29/2021.
We retain tweets whose texts tend to be argumentative, with an argument confidence score larger than 0.9 by using ArgumentText Classify API\footnote{https://api.argumentsearch.com}. 99.48\% of tweets are discarded for having an argument confidence score below 0.9. These filtering processes ensure our annotation data has high argumentation-confidence and topic-relevance.
\subsection{Annotation Strategies}
\label{Sec: annotation_strategies}
\begin{table}[t]\small
\centering
\begin{tabular}{lcc}
\hline
Task & Alpha & Count\\
\hline 
Stance & 64.5 & 87\\
% persuasiveness improvement & 0.47 & 51\\
% persuasiveness level & 0.19 & 38\\
Content type & 71.1 & 38\\
Persuasion mode& 19.9 & 38\\\hline
\end{tabular}
\caption{First pilot annotation inter-agreement on \textit{gun control} topic. Persuasion modes are annotated as single choices from logos, pathos, and ethos.}\label{tab:agreement_pilot}
\end{table}

\begin{table}[t]\small
\centering
\begin{tabular}{lcc}
\hline
Task & Alpha & Count\\
\hline
Stance & 76.1 & 1003\\
Persuasiveness* & / & 1003 \\
Content type & 64.6 & 1003\\
Logos & 55.3 & 259\\
Pathos & 51.0 & 259\\
Ethos & 57.8 & 259\\\hline
\end{tabular}
\caption{Inter-agreement rate of each annotation task in our final corpus on \textit{gun control} topic, and the number of samples with the corresponding annotation. (*) We only show numbers of persuasiveness since they are annotated with average persuasiveness scores from annotators rather than labels. 
% We use and threshold ($\gamma$) rather than agreed by annotators.
}
\label{tab:agreement_mturk}
\end{table}

\begin{figure}[t]
    \centering
    \includegraphics[width=1\columnwidth]{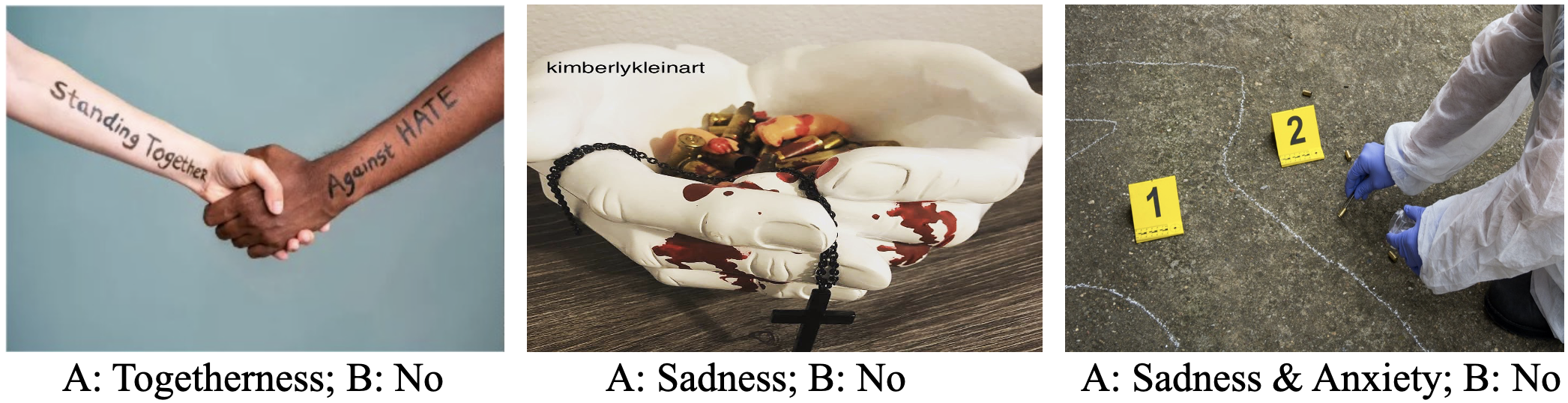}
    \caption{Annotator A annotates the above images as Pathos because these examples express emotions, while annotator B disagrees and marks as not Pathos.}
    \label{fig:example_pathos}
\end{figure}

We develop annotation strategies based on several rounds of pilot annotations. To ensure the annotation quality, we provide coding manual and examples for annotators 
(see the Appendix \ref{sec: coding manual appendix} for details). 
We employ qualified workers who passed a qualitative test that evaluates the workers' understanding on our annotation manual.

We start with the topic of \textit{gun control}. In the first-round, we distribute 87 samples to two random annotators on MTurk. Table \ref{tab:agreement_pilot} shows Krippendorff's alpha \cite{krippendorff2011computing} score for inter-rater agreement\footnote{Note that the availability of annotation questions is based on the answer to the prior questions (Fig. \ref{fig:pipeline})
% (e.g., if tweet posts irrelevant topic, was asked questions about persuasiveness will not be asked), 
therefore each task has different sample numbers.}.
%  and count numbers. Note that there is no inter-rater agreement on the persuasiveness score as they are quantitative results.
% the threshold decides the results.
Based on the interpretation of alpha scores in \citet{landis1977measurement,hartling2012validity}, we conclude that stance and
% , which indicates that the viewpoints within the sampled tweets are highly polarized and clearly expressed.
% Annotators reach a moderate agreement on whether the image improved the persuasiveness. However, the agreement on the level of persuasiveness (Low, including L1, L2; High, including L3, L4, L5 defined in Sec. \ref{Sec: Score}) is slight. One reason would be that differentiating if the images are highly helpful or less helpful is relatively subjective.
content type have a substantial inter-agreement but persuasion mode inter-agreement is slight.  
% Concerning the coverage of our defined scheme, there is no case where the persuasion mode is classified as "other" or the content type is not included in our six categories. 
To investigate this issue, we modify our annotation guideline for persuasion mode. Instead of using three-class annotation (i.e., choosing one persuasion mode from 3 options), we move to three-label annotation that asks a binary question for each mode for each sample (i.e., annotating yes/no for each persuasion mode, individually). 
% Specifically, the annotator is asked to decide whether the image appeals to each persuasion mode individually.
Moreover, the annotators are required to justify their choices by giving short comments.
% Second, we assign three annotators for each sample instead of two to prevent biases.
% In addition, 
% we improve the annotation quality by \textbf{(1)} adding direct definitions to the persuasion mode and \textbf{(2)} 
% we develop qualification exams for annotators,
% only permitting annotators that get high scores in qualification tests and \textbf{(3)} asking for reasons of choices. 
% The results of the persuasion mode study
The improved results (on the final corpus from Sec.~\ref{Sec: statistics}) are shown in Table \ref{tab:agreement_mturk}, although the persuasion mode agreement (i.e., Logos, Pathos, and Ethos) is still lower than %To this end,\ 
stance and content type.  %receive substantial agreement, but the persuasion modes are less agreed. 
% logos and ethos receive moderate agreement, yet pathos is fair.
% Observations on disagreement cases show that annotators understand the coding manual about what to do. However,
This is likely because annotators have different emotional reasoning (i.e., some annotators are easily evoked by images while others are not).
% (we may not even agree with each other on those samples). It is not surprising to find out that identifying emotion incitement
% is more subjective than logic and credibility. 
For example,  %annotators label different modes for the same image. For example, 
one annotator recognized strong emotional impact (e.g., togetherness, sadness, anxiety, etc.), while the other not as shown in Fig. \ref{fig:example_pathos}. 

% The observations demonstrate that one image can potentially show different/multiple persuasion modes to audiences. And using binary annotation on each mode instead of exclusive choosing one mode leads to a better agreement on this subjective task.

\begin{figure}[htbp]
    \centering
    \includegraphics[width=1\columnwidth]{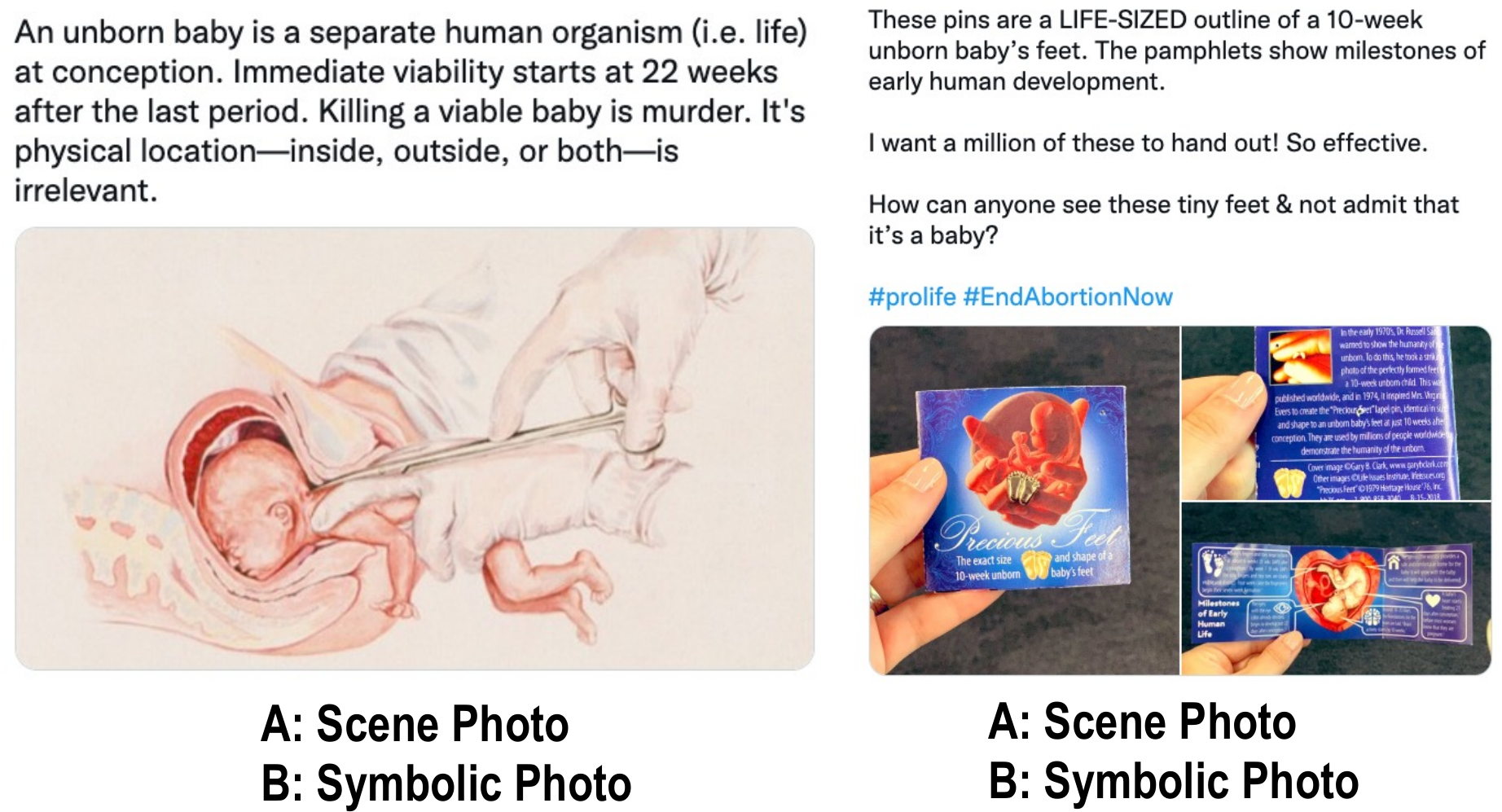}
    \caption{Samples of disagreed on the content type in the topic \textit{abortion}.}
    \label{fig:example_abortion}
\end{figure}

% We conduct the third round of pilot study on gun control on 100 more samples to comprehensively test our updated annotation scheme. 
% In our third-round annotation, we 
% The results are shown in Table \ref{tab:agreement_mturk}. 
% The pilot studies on the other two topics also show a similar pattern: While the stance, content type, and persuasiveness receive substantial agreement, the persuasion modes are less agreed upon. 
% After annotating \textit{gun control},  we also perform pilot annotations for the \textit{immigration} and \textit{abortion} topics.  
We further perform pilot annotations for the topics of \textit{immigration} and \textit{abortion}, with the best annotation strategies that we developed for annotating \textit{gun control}.
% These receive substantial/moderate agreement on stance and content type but pathos is only fair (details are described in Appendix \ref{sec: pilot appendix}).
We randomly choose 100 or 200 tweets respectively on \textit{immigration} or \textit{abortion} for the pilot study, and make a topic-specific instruction for the stance annotation that provides some topic-specific examples. 
%For the stance and content type, we assign two annotators for each sample; for the persuasion mode, we assign three annotators for each sample. We focus on the most challenging part learned from the pilot study for the "gun control" topic. Specifically, instead of studying the Kappa score on the persuasion mode of images, we directly calculate Krippendorff's alpha scores for three image persuasion modes. 
The Inter-rater Agreement for both topics is shown in Table \ref{tab:pilot_immigation_abortion}. We observe high Inter-rater Agreements on the stance annotation, which demonstrates the utility of our topic-specific instructions. The agreement on the content type is generally good, however, \textit{abortion} has relatively lower agreement than the other two topics. One main reason is that authors prefer using photos to support their arguments. Such photos lead to ambiguity between scene photos and symbolic photos, as examples shown in Fig. \ref{fig:example_abortion}. Moreover, we notice that the agreements on the persuasion modes are not satisfying. For \textit{immigration}, Ethos has the lowest agreement, and one explanation is that there are few authentic resources that provide credible and trustworthy arguments on this topic; for \textit{abortion}, the agreement on all three persuasion modes are relatively low, in particular, Logos surprisingly gets the lowest agreement. 
%We plan to investigate these relationships between topics and persuasion modes in our future study. This indicates that the different topics might potentially prefer different persuasion modes.

These studies indicate that the inter-rater agreement on annotating persuasion mode is topic-dependent, and the relationship between topics and persuasion modes needs further investigation. We thus create the first version of \textit{ImageArg} data using only the \textit{gun control} topic, and leave the other two topics for future work. 

% \begin{table}[htbp]\small
% \centering
% \begin{tabular}{lcc}
% \hline
% Task & Alpha & Count\\
% \hline 
% stance & 61.5 & 100\\

% content type & 65.8& 53\\
% logos & 56.7 & 23\\
% pathos & 46.0 & 23\\
% ethos & 30.8 & 23\\\hline
% \end{tabular}
% \caption{Inter-agreement rate of each annotation task on the topic "immigration." The count represents the number of samples after filtering from previous questions.}\label{tab:pilot_immigation}
% \end{table}

\begin{table}[t]\small
\centering
\begin{tabular}{lcccc} \hline
\multicolumn{1}{c}{\multirow{2}{*}{Task}} & \multicolumn{2}{c}{Immigration} & \multicolumn{2}{c}{Abortion} \\ \cline{2-5} 
\multicolumn{1}{c}{}                      & Alpha          & Count          & Alpha         & Count        \\ \hline
Stance                                    & 61.5           & 100            & 68.7          & 200          \\
Content type                              & 65.8           & 53             & 56.6          & 76           \\
Logos                                     & 56.7           & 23             & 25.0          & 48           \\
Pathos                                    & 46.0           & 23             & 37.5          & 48           \\
Ethos                                     & 30.8           & 23             & 28.2          & 48           \\ \cline{1-5}
\end{tabular}
\caption{Inter-agreement rate of each annotation task on the topic \textit{immigration} and \textit{abortion}. The count represents the number of samples after filtering from previous questions.}\label{tab:pilot_immigation_abortion}
\end{table}

% However, the agreements on the persuasion modes are still low. Unlike "gun control," which receives the lowest agreement on Pathos, the agreement on Ethos is the lowest in "immigration." One potential reason is that there are not many resources that provide credible and trustworthy resources for the arguments on the topic. This indicates that the different topics might potentially prefer different persuasion modes.

% \begin{table}[htbp]\small
% \centering
% \begin{tabular}{lcc}
% \hline
% Task & Alpha & Count\\
% \hline 
% stance & 68.7 & 200\\

% content type & 56.6 & 76\\
% logos & 25.0 & 48\\
% pathos & 37.5 & 48\\
% ethos & 28.2 & 48\\ \hline
% \end{tabular}
% \caption{Inter-agreement rate of each annotation task on the topic "abortion." The count represents the number of samples after filtering from previous questions.}\label{tab:pilot_abortion}
% \end{table}

% To this end, we change our scheme of annotating persuasion mode.

% With respect to the coverage of our defined scheme, there is no case where the persuasion mode is classified as \emph{Other} or where the content type is not included in our six categories. 

% \begin{table}[t]
% \centering
% \begin{tabular}{p{1.5cm} p{1cm} p{1cm} p{1cm} p{1cm}}
% \hline
%      &Logos & Pathos & Ethos & Other\\ \hline
% Logos & 8 & 1 & 4 & 0 \\\hline
% Pathos & 6 & 8 & 4 & 0 \\\hline
% Ethos & 3 & 2 & 2 & 0\\\hline
% Other& 0 & 0 & 0 & 0\\ \hline
% \end{tabular}
% \caption{Confusion matrix of persuasion mode annotation between annotator 1 and annotator 2.}\label{tab:cm_role_mturk}
% \end{table}
\begin{figure*}[t]
    \centering
    \includegraphics[width=1.2\columnwidth]{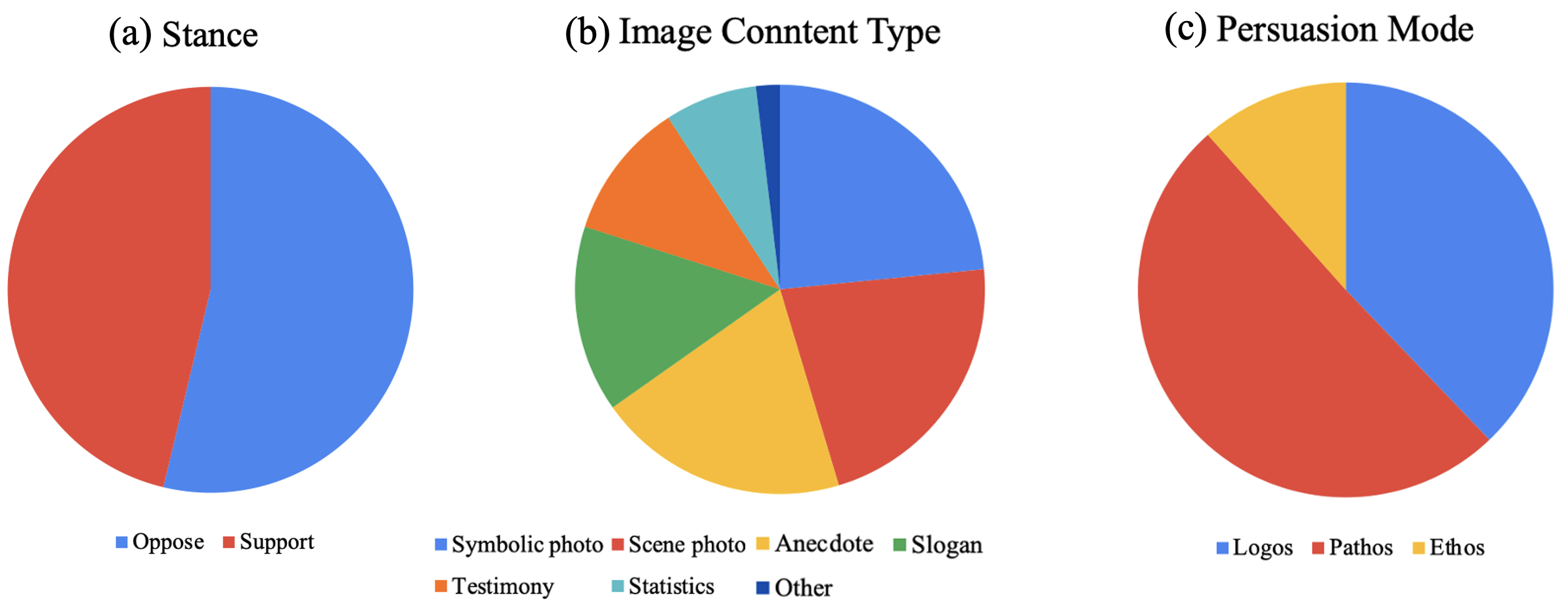}
    \caption{Distributions of (a) stance, (b) image content type, and (c) persuasion mode in our corpus on \textit{gun control} topic.}
    \label{fig:image-dist}
\end{figure*}
\begin{figure*}[t]
    \centering
    \includegraphics[width=1.6\columnwidth]{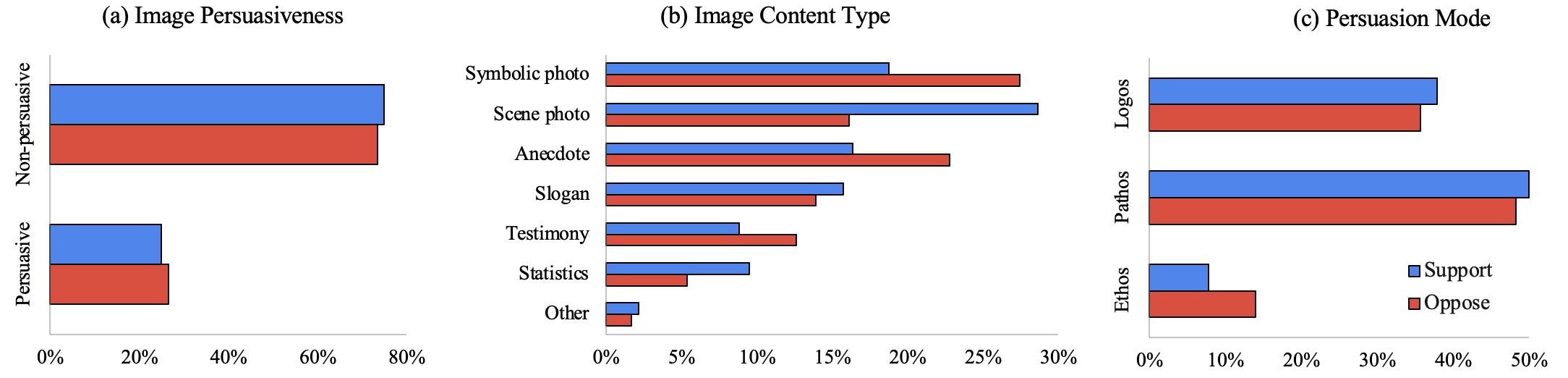}
    \caption{Distributions of (a) image persuasiveness, (b) content type and (c) persuasion mode regarding stances (support in blue and oppose in red) in our corpus on \textit{gun control} topic.}
    \label{fig:image-stances}
\end{figure*}

\begin{figure*}[t]
    \centering
    \includegraphics[width=1.2\columnwidth]{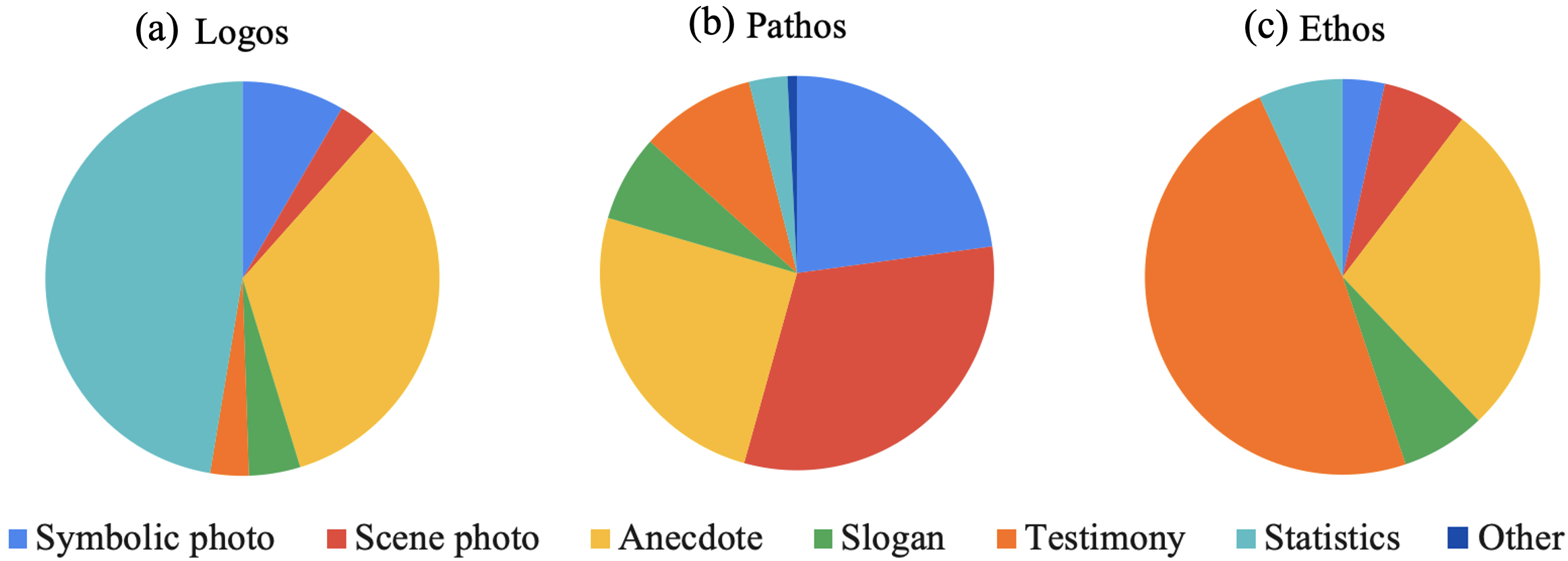}
    \caption{Distributions of image content type in different persuasion mode (a) Logos, (b) Pathos, and (c) Ethos in our corpus on \textit{gun control} topic.}
    \label{fig:image-type}
\end{figure*}

% \section{Annotation Analysis}

% \section{Annotation Analysis}
% We introduce the pilot study. We evaluate the reliability of our annotation scheme with inter-rater agreement using Cohen's kappa, Krippendorff’s alpha \cite{krippendorff2011computing} and Fleiss’ Kappa\cite{falotico2015fleiss} scores.  
% Since the availability of some questions is based on the answer to the prior questions (e.g. if the tweet is not related to the topic then questions about persuasiveness won't be asked), we also report the number of samples to be evaluated for each annotation task. 
\subsection{Corpus Statistics and Analysis}
\label{Sec: statistics}

\begin{table}[t]\small
\centering
\begin{tabular}{ccc}\hline
Persuasiveness Score & Count & Percentage \\
\hline
0.0 - 0.1           & 336   & 33.50\%     \\
0.1 - 0.3           & 232   & 23.13\%    \\
0.3 - 0.5           & 176   & 17.55\%    \\
0.5 - 0.7           & 118   & 11.76\%    \\
0.7 - 0.9           & 66   & 6.58\%    \\
$\geq$ 0.9       & 75    & 7.48\%   \\
\hline
\end{tabular}
\caption{The annotated image persuasiveness score distribution on
\textit{gun control} topic in \textit{ImageArg}.}
\label{tab: persuasiveness-stats}
\end{table}

We annotate 1003 samples that hold a support or oppose stance on \textit{gun control} topic. 36\% of data is discarded for not having an agreed support/oppose stance. We report the distribution of each annotation scheme in Fig. \ref{fig:image-dist}, and the inter-rater agreement evaluation in Table \ref{tab:agreement_mturk}.
The results reveal that the annotators have substantial agreement on the stance and content types, and moderate agreement on the image persuasion mode. Specifically, the stance annotations are balanced distributed as shown in Fig. \ref{fig:image-dist} (a): 46.3\% support and 54.7\% oppose. As for image persuasiveness annotations,
%Regarding image persuasiveness, we focus on the persuasiveness gains when an image is added to a text tweet. As described in Sec. \ref{Sec: Score}, we annotate image persuasiveness scores. 
Table \ref{tab: persuasiveness-stats} shows sample distributions in different persuasiveness score intervals. We use a threshold $\gamma$ to discretize numerical persuasiveness scores to binary labels (i.e., persuasiveness or not). The $\gamma$ is set to 0.5 in our annotations since the persuasiveness score is an average of three annotators, thus $\gamma$ greater than 0.5 suggests that there is at least two annotators annotating images persuasiveness with L1 or higher ($\geq1$) scores (as defined in Sec. \ref{Sec: Score}) or at least one annotator annotating L2 or higher scores ($\geq$ 2).
% (259, 25.8\% negative and 744, 74.2\% positive).
In terms of image content types, its distribution is shown in Fig. \ref{fig:image-dist} (b): Symbolic photo (23.43\%), Scene photo (21.93\%), Anecdote (19.84\%), Slogan (14.76\%), Testimony (10.87\%), Statistics (7.28\%), Other (1.89\%). We observe that images (i.e., symbolic photo/scene photo) occupy a high proportion of the samples, in contrast, data evidence (i.e., statistics) takes the relatively low ratio. One potential reason is that social media contents like tweets are generally short and informal, which prefers relatively simple evidence. Note that there are 19 ``other" out of 1003 annotations that annotators were confused about; however, it does suggest that our image content type scheme works very well as only 1.89\% are out of our defined labels. In terms of image persuasion mode, we only annotate images with persuasiveness score $\gamma$ greater than 0.5, which produces 259 samples. As shown in Fig. \ref{fig:image-dist} (c), we have 37.85\% Logos, 50.60\% Pathos, and 11.55\% Ethos.

Additionally, we show how the stance impacts image persuasiveness, content type, and persuasion mode. In Fig. \ref{fig:image-stances} (a), supporting and opposing \textit{gun control} stance are almost evenly distributed with respect to persuasiveness and non-persuasiveness, which suggests that images generally support both positive and negative arguments.
% The distributions of different dimensions in terms of stances are shown in Figure . 
For the image content type in Fig. \ref{fig:image-stances} (b), opposing \textit{gun control} stance uses significantly more images with respect to Symbolic photos, Anecdote, and Testimony; however, supporting stance prefers images in the content of Scene photos and Statistics.
Regarding persuasion mode in Fig. \ref{fig:image-stances} (c), images in supporting \textit{gun control} stance uses more Logos and Pathos but less Ethos than those in the opposing stance. 
%This implies that better persuasiveness can be achieved with a dedicated selected persuasion mode.

To further study the relevance between image content type and persuasion mode, we report their  correlated distributions in charts. Fig. \ref{fig:image-type} (a) shows that most Logos samples use Statistics and Anecdote evidence. It meets the intuition that the logical reasoning can usually be clarified by introducing anecdotes and justified by providing supportive statistics. 
In terms of Pathos in Fig. \ref{fig:image-type} (b), the majority of samples utilize Scene and Symbolic photos. This is also reasonable since images generally promote emotional impression by presenting visual information.
Regarding Ethos, Fig. \ref{fig:image-type} (c) shows Testimony takes the most ratio because statements from authorities can enhance trustworthiness. 
These correlations imply mutual influences between different annotation dimensions and raise demands for further study.

\section{Experiments}
\subsection{Models and Tasks}
We evaluate our corpus on \textit{gun control} topic with binary classification tasks for Stance, Persuasiveness, Logos, Pathos, and Ethos and multi-class classification task for Image Content. 
Since data size is relatively small, we use pretrained image encoder ResNet50 \cite{he2016deep}
% ,  VGG \cite{simonyan2014very}
 and text encoder BERT \cite{devlin2019bert} to fine-tune linear classifiers. For fair comparison, we project both image and text embeddings into 1024 dimension before feeding into classification layers. We compare task performance on Text Modality (T-M), Image Modality (I-M), and Image-Text Multi-modality (M-M) that concatenates T-M and I-M. As for baseline (BASE), we report the performance when all samples are predicted as positive for binary classification, or predicted as the majority label for multi-class classification. We don't use the majority baseline for the binary classification task because the recall and F1 scores are always 0 if the majority label is negative, which is not interesting to compare with.

In the implementation, we follow the annotation strategy (Sec. \ref{Sec: annotation_strategies}) that uses threshold $\gamma$ equal to 0.5 to encode persuasiveness scores into binaries.
% Since persuasion mode annotated with multi-labels (e.g., an example can be labeled with both logos and pathos), we did multi-label classification for all 259 persuasion mode and binary classification for its subcategories (logos, pathos, and ethos). 
We remove Emoji, URLs, Mentions, and Hashtags in tweet texts, and discard 19 samples labeled with ``Other" for the image content classification task. All images are resized to 224×224 dimension, and augmented (i.e., horizontal-flipped) only in training. Our models are implemented with Pytorch, and trained on a GeForce RTX 3080 GPU. We freeze BERT and ResNet50 encoders while training classifiers, and optimize the networks using Adam optimizer with $\beta_1=0.9, \beta_2=0.999, \epsilon=10^{-8}$. The learning rate is 0.001 and the batch size is 16. We conduct 5-fold cross-validation (80\% data in train; 20\% data in test). We report 5-fold average Precision, Recall, F1, and AUC scores for binary classification and macro Precision, Recall, and F1 scores for multi-class classification on the test set.
%We use positive-label predictions as our baseline for binary classification, and majority-label prediction for multi-class baseline. 
\subsection{Quantitative Results Analysis}
\label{Sec: quan_experiments}

\begin{table}[t]
\centering
\begin{adjustbox}{width=1\columnwidth}
\begin{tabular}{cccccc}
\hline
Task & Model & Precision & Recall & F1 & AUC\\
\hline
\multirow{4}{*}{\begin{tabular}[c]{@{}c@{}}Stance\\ (binary)\end{tabular}}   
& BASE    & $0.470_{\pm0.02}$ & $\mathbf{1.000_{\pm0.00}}$ & $\mathbf{0.639_{\pm0.02}}$ & / \\
 & T-M       & $\mathbf{0.501_{\pm0.05}}$ & $0.740_{\pm0.03}$ & $0.596_{\pm0.04}$ & $\mathbf{0.527_{\pm0.04}}$  \\
 & I-M      & $0.443_{\pm0.08}$ & $0.147_{\pm0.03}$ & $0.218_{\pm0.04}$ & $0.472_{\pm0.05}$  \\
 & M-M  & $0.414_{\pm0.04}$ & $0.369_{\pm0.06}$ & $0.390_{\pm0.05}$ & $0.417_{\pm0.03}$ \\
 \hline
\multirow{4}{*}{\begin{tabular}[c]{@{}c@{}}Persua.\\ (binary)\end{tabular}} 
 & BASE   & $0.257_{\pm0.03}$ & $\mathbf{1.000_{\pm0.00}}$ & $\mathbf{0.408_{\pm0.04}}$ & / \\
 & T-M         & $0.260_{\pm0.01}$ & $0.725_{\pm0.11}$ & $0.380_{\pm0.01}$ & $0.502_{\pm0.03}$ \\
 & I-M       & $\mathbf{0.313_{\pm0.02}}$ & $0.196_{\pm0.05}$ & $0.238_{\pm0.05}$ & $0.528_{\pm0.03}$ \\
 & M-M  & $0.296_{\pm0.05}$ & $0.486_{\pm0.05}$ & $0.364_{\pm0.03}$ & $\mathbf{0.534_{\pm0.04}}$ \\
  \hline
   \multirow{4}{*}{\begin{tabular}[c]{@{}c@{}}Content\\ (6-class)\end{tabular}}
 & BASE  & $0.041_{\pm0.00}$ & $0.167_{\pm0.00}$ & $0.066_{\pm0.00}$ & /\\
 & T-M      & $0.198_{\pm0.08}$ & $0.201_{\pm0.03}$ & $\mathbf{0.165_{\pm0.03}}$ & /\\
 & I-M     & $\mathbf{0.235_{\pm0.09}}$ & $\mathbf{0.204_{\pm0.02}}$ & $0.151_{\pm0.02}$ & / \\
 & M-M  & $0.200_{\pm0.02}$ & $0.179_{\pm0.01}$ & $\mathbf{0.165_{\pm0.01}}$ & / \\
 \hline
\multirow{4}{*}{\begin{tabular}[c]{@{}c@{}}Logos\\ (binary)\end{tabular}} 
& BASE   & $\mathbf{0.405_{\pm0.05}}$ & $\mathbf{1.000_{\pm0.00}}$ & $\mathbf{0.575_{\pm0.05}}$ & / \\
 & T-M      & $0.364_{\pm0.08}$ & $0.613_{\pm0.13}$ & $0.456_{\pm0.10}$ & $0.439_{\pm0.08}$ \\
 & I-M     & $0.351_{\pm0.22}$ & $0.097_{\pm0.07}$ & $0.144_{\pm0.10}$ & $0.406_{\pm0.08}$ \\
 & M-M  & $0.262_{\pm0.27}$ & $0.047_{\pm0.05}$ & $0.077_{\pm0.08}$ & $\mathbf{0.508_{\pm0.06}}$ \\
 \hline
\multirow{4}{*}{\begin{tabular}[c]{@{}c@{}}Pathos\\ (binary)\end{tabular}}   
& BASE & $0.554_{\pm0.04}$ & $\mathbf{1.000_{\pm0.00}}$ & $\mathbf{0.712_{\pm0.04}}$ & / \\
 & T-M        & $0.613_{\pm0.11}$ & $0.714_{\pm0.08}$ & $0.658_{\pm0.09}$ & $0.582_{\pm0.10}$ \\ 
 & I-M       & $\mathbf{0.666_{\pm0.09}}$ & $0.184_{\pm0.07}$ & $0.280_{\pm0.07}$ & $\mathbf{0.593_{\pm0.09}}$ \\
 & M-M  & $0.471_{\pm0.42}$ & $0.071_{\pm0.10}$ & $0.114_{\pm0.15}$ & $0.507_{\pm0.12}$ \\
 \hline
\multirow{4}{*}{\begin{tabular}[c]{@{}c@{}}Ethos\\ (binary)\end{tabular}} 
& BASE & $0.128_{\pm0.04}$ & $\mathbf{1.000_{\pm0.00}}$ & $0.226_{\pm0.06}$  & / \\
 & T-M        & $0.168_{\pm0.05}$ & $0.817_{\pm0.15}$ & $\mathbf{0.272_{\pm0.06}}$ & $\mathbf{0.580_{\pm0.09}}$ \\
 & I-M       & $\mathbf{0.244_{\pm0.16}}$ & $0.233_{\pm0.16}$ & $0.221_{\pm0.13}$ & $0.459_{\pm0.18}$ \\
 & M-M  & $0.124_{\pm0.15}$ & $0.083_{\pm0.11}$ & $0.098_{\pm0.12}$ & $0.450_{\pm0.09}$ \\  
 \hline

 \hline
\end{tabular}
\end{adjustbox}
\caption{Classification benchmark results with standard deviation on \textit{gun control} topic in \textit{ImageArg} corpus. Note that the reported Persuasiveness results use threshold $\gamma$ equal to 0.5. The Stance, Persuasiveness, and Image Content tasks use 1003 annotations; The Logos, Pathos, and Ethos use 259 annotations.}
\label{tab:binary}
\end{table}

% \begin{table}[t]
% \begin{adjustbox}{width=1\columnwidth}
% \begin{tabular}{llccc}
% \hline
% Data & Encoder & Precision & Recall & F1 \\
%  \hline
%  text  & BERT           &50.89&    50.95&   50.06 \\
%  image &VGG16       &28.49&    50.00&   36.23\\
%  image &ResNet50  &53.78&    52.62&   49.82 \\
%  image &ResNet101 &54.68&    \textbf{53.81}&   52.31 \\
%  img+text &VGG16+BERT       &\textbf{54.79}&    53.62&   48.45 \\
%  img+text &ResNet50+BERT  &53.65&    53.67&   \textbf{53.56} \\
%  img+text &ResNet101+BERT &49.46&    49.59&   48.39 \\
%  \hline
% \end{tabular}
% \end{adjustbox}
% \caption{Persuasiveness classification with BERT-large text encoder, VGG16, ResNet50, ResNet101 image encoder, and persuasiveness score threshold $\gamma=0.3$.}
% \label{tab:persuasion_mode_threshold}
% \end{table}

% Please add the following required packages to your document preamble:
% \usepackage{multirow}
\begin{table}[t]
\centering
\begin{adjustbox}{width=1\columnwidth}
\begin{tabular}{ccccc}
\hline
\multirow{2}{*}{\begin{tabular}[c]{@{}c@{}}Threshold \\ ($\gamma$)\end{tabular}} & \multirow{2}{*}{Pos. Ratio} & \multicolumn{3}{c}{F1 Score} \\ \cline{3-5} 
 &  & T-M & I-M & M-M \\ \hline
0.1 & 66.50\% & $\mathbf{0.681_{\pm0.02}}$ & $\mathbf{0.265_{\pm0.05}}$ & $\mathbf{0.536_{\pm0.03}}$ \\
0.3 & 43.37\% & $0.538_{\pm0.03}$ & $0.251_{\pm0.04}$ & $0.459_{\pm0.05}$ \\
0.5 & 25.8\% & $0.380_{\pm0.01}$ & $0.238_{\pm0.05}$ & $0.364_{\pm0.03}$ \\
0.7 & 14.1\% & $0.246_{\pm0.02}$ & $0.168_{\pm0.04}$ & $0.233_{\pm0.01}$ \\
0.9 & 7.48\% & $0.138_{\pm0.03}$ & $0.084_{\pm0.03}$ & $0.115_{\pm0.01}$ \\ \hline
\end{tabular}
\end{adjustbox}
\caption{F1 scores with standard deviation and positive label ratio for Persuasiveness classification with respect to different threshold ($\gamma$).}
\label{tab:persuasion_mode_threshold}
\end{table}

% \begin{table}[t]
% \centering
% \begin{adjustbox}{width=0.8\columnwidth}
% \begin{tabular}{lccc}
% \hline
%  Model & Precision & Recall & F1 \\
%  \hline
% % \multirow{3}{*}{mode} 
%   Baseline    & 9.22 & \textbf{19.33} & \textbf{12.45} \\
%   Text        & 5.91 & 7.5 & 3.06  \\
%   Image       & \textbf{11.01} & 8.82 & 8.16 \\
%   Img+Text     & 9.91 & 14.74 & 7.85  \\
%  \hline
% \end{tabular}
% \end{adjustbox}
% \caption{Multi-label classification (persuasion mode) benchmark results on \textit{ImageArg} dataset.}
% \label{tab:multilabel}
% \end{table}

% \begin{table}[t]
% \centering
% \begin{adjustbox}{width=0.8\columnwidth}
% \begin{tabular}{lccc}
% \hline
%  Model & Precision & Recall & F1 \\
%  \hline
% % \multirow{3}{*}{content}
%  Baseline  & 4.09 & 16.67 & 6.57 \\
%  Text      & 19.81 & \textbf{20.06} & \textbf{16.50} \\
%  Image     & \textbf{23.50} & 20.4 & 15.08 \\
%  Img+Text  & 19.96 & 17.93 & 16.45 \\
%  \hline
% \end{tabular}
% \end{adjustbox}
% \caption{Multi-class classification (image content) benchmark results on \textit{ImageArg} dataset.}
% \label{tab:multiclass}
% \end{table}

Table \ref{tab:binary} shows the classification benchmark results with standard deviation on \textit{gun control} topic in \textit{ImageArg} corpus. 
% that the persuasiveness and persuasion mode gain significant improvement if leverage image modality, which demonstrates that images are significant modality that can contributes to the persuasiveness of linguistic arguments in social content.

\textbf{Task-Stance} Regarding stance, T-M has the highest performance in terms of AUC scores. It reveals that the image information is redundant to the text for identifying the stance; moreover, the image might introduce disturbing noise due to limited training samples.

%Our reasoning is that text conveys more explicit information compared to images, in contrast, additional image modality may make stance implicit. 
\textbf{Task-Persuasiveness} As for persuasiveness task, we observe that  M-M performs slightly poorer than T-M regarding F1 score but relatively better in AUC score. This is because persuasiveness (positive/negative) labels are unbalanced if we use $\gamma=0.5$ (as shown in Table \ref{tab: persuasiveness-stats}). We show F1 scores drop with respect to threshold increases from 0.1 to 0.9 in Table \ref{tab:persuasion_mode_threshold}. 

%This suggests that small persuasiveness score would help distinguish persuasiveness and non-persuasiveness, which would help persuasiveness score annotation for "\textit{immigration}" and "abortion" topics in our future work. 
\textbf{Task-Content} 
% Table \ref{tab:binary} shows
In terms of 6-class classification for image content, although all modalities outperform the baseline, the task is shown to be very challenging. It is surprising that the performance with I-M is lower than T-M. The reason might be that visual argumentative tasks demand more specific image encoders that learn sufficient knowledge on persuasiveness and social science; however, the used image encoder is pretrained on a general object detection task on the ImageNet \cite{krizhevsky2012imagenet}, thus our model is unable to learn well for this argumentative task with very limited training data. 

\textbf{Task-Logos} Regarding logos, we observe that M-M gains the best AUC score but I-M has lower AUC than T-M. The reason might be that logos images usually contain statistic charts, as shown in Fig. \ref{fig:image-type} (a), that are relatively more difficult to encode than normal images (e.g., images with explicit objects), but multi-modal models might learn these patterns directly from textual inputs.

%Another unusual observation is that multimodality has lowest F1 score. This indicates the multimodality model did not learn well as it uses small size (98 positive and 161 negative logos) annotations, which is very unbalanced and extremely difficult for training a good multi-modality model. 
\textbf{Task-Pathos} As for pathos, I-M has the best performance in terms of AUC score, and T-M is quite close to I-M while M-M has the lowest. This suggests that the multi-modal representation fusion method we used might be too weak to conduct complex reasoning on the pathos task.

%image modality has slight improvement in terms of precision and AUC scores but drops dramatically on recall and f1 scores even if data is balanced (131 positive and 128 negative pathos). 
%This suggests that pre-trained image model is able to recognize some images with high precision, but failed to learn well on other images such as anecdote that are text overlay images (as shown in Figure \ref{fig:image-role}). 
\textbf{Task-Ethos} The best performance in ethos is from T-M. It is intuitive because the image encoder pre-trained on object detection is unable to recognize the optical characters on the image, while this kind of images are common in ethos, e.g., testimony images in Fig. \ref{fig:image-type} (c).

\subsection{Qualitative Results Analysis}
\label{Sec: qual_experiments}
\begin{figure}[t]
    \centering
    \includegraphics[width=1\columnwidth]{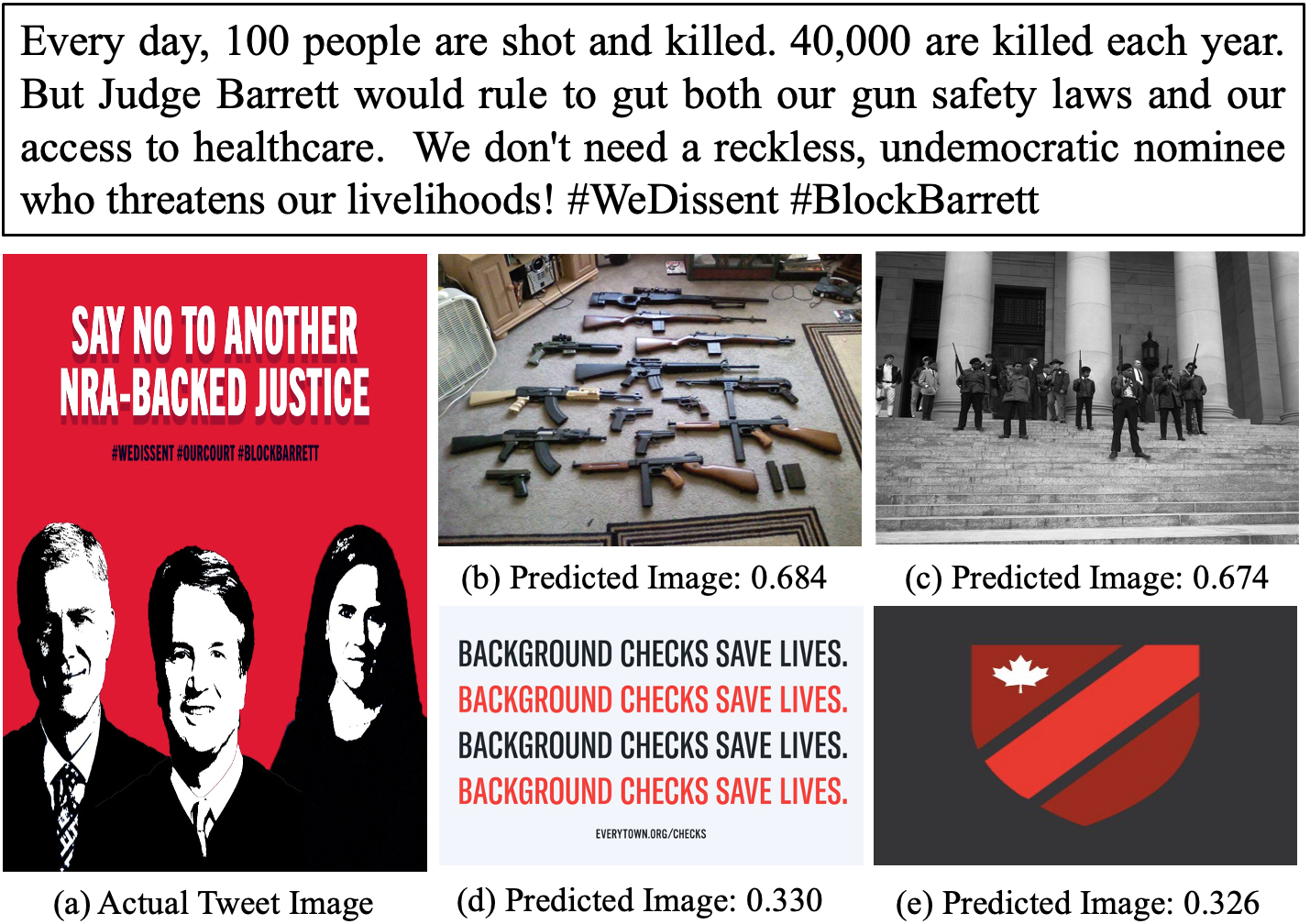}
    \caption{(a) the actual tweet image annotated with persuasiveness score 0 in \textit{ImageArg}; (b) and (c) with top predicted persuasiveness scores; (d) and (e) with lowest predicted persuasiveness scores while retrieving images given the same tweet text.}
    \label{fig:image-example-result}
\end{figure}

We conduct qualitative analysis by retrieving the most persuasive images given a text. Specifically, we run the multi-modality (M-M) model, trained for the persuasiveness task, on the test set in each fold (out of 5 folds). The inputs are image-text pairs of which all candidate images are paired with the same text, and the outputs are image persuasiveness scores. Fig. \ref{fig:image-example-result} shows the actual, top, and bottom images with the highest and lowest persuasiveness scores, respectively. It is interesting to find that images with specific objects or scenes (image (b), and (c) in Fig. \ref{fig:image-example-result}) boost the persuasiveness scores; however, images with slogans or symbolism have lower scores  (image (d), and (e) in Fig. \ref{fig:image-example-result}). This suggests that our image encoder is capable of capturing object information but not optical characters on images (e.g., slogans); therefore, our retrieved images with best persuasion scores are mostly related to gun-object images. Thus, learning an image encoder pre-trained on slogans and visual symbolism is a promising future direction to improve the performance. In the meanwhile, extracting text information from images by OCR tools and use it as an auxiliary modality may help models learn the context.

\section{Conclusion and Future Work}
We create a brand-new multi-modal persuasiveness dataset \textit{ImageArg} that focuses on image functionality and persuasion mode for persuasive arguments. We extend the argumentative annotation scheme from text to vision, and demonstrate its feasibility. We then establish a benchmark on our defined tasks using computational models, with multiple input modalities. Our experimental results reveal that image persuasiveness mining is challenging and that there is ample room for model improvement. We identify the image encoder as a key modeling bottleneck through a series of qualitative and quantitative analysis, which offers a good starting point for further exploration on this rich and challenging topic. The first version of \textit{ImageArg} has 1003 annotations on the \textit{gun control} topic. In the future work, we will work on constructing datasets on the topics of \textit{immigration} and \textit{abortion}, and scaling up the annotations.
%We benchmarked the dataset to demonstrate the image modality as an argumentation unit shows potentials to improve persuasiveness of linguistic arguments in tweets. Our proposed annotation scheme is demonstrated effective through our qualitative and quantitative analysis for the gun control topic. However, the inter-agreement of the persuasion mode is not good enough for the \textit{abortion} and \textit{immigration} topics as shown in Appendix \ref{sec: pilot appendix}. We leave these two topics as our future direction. 

\bibliography{anthology,acl2020}
\bibliographystyle{acl_natbib}

\newpage
\clearpage
\appendix

\section{Coding Manual}
\label{sec: coding manual appendix}
\subsection{Stance} \label{Sec: stance code manual}
We setup different instructions for stance annotations on different topics since we would like to provide detailed instructions and examples for different topics separately.
\subsubsection{Stance: Gun Control}
\begin{figure}[htbp]
    \centering
    \includegraphics[width=0.7\columnwidth]{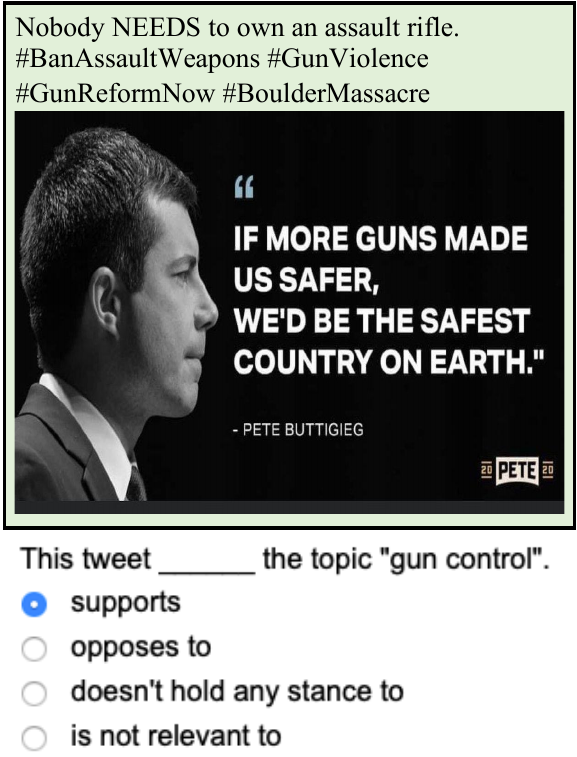}
    \caption{Example of stance annotation on gun control}
    \label{fig:stance_guncontrol}
\end{figure}

We aim to study the topic and the stance of tweets. Given a tweet accompanying with an image, you need to answer the stance of the tweet towards a given topic, as depicted in Figure \ref{fig:stance_guncontrol}.
Please make sure that you have the basic knowledge about that social topic and you understand the key message that the tweet (i.e. both the text and the image) sends. Just skip the HIT if you are not sure.

The question is about the stance. You need to decide whether the tweet is relevant or not to the social topic gun control. If it is relevant, then you need to annotate the stance: supports/opposes to/doesn’t hold any stance. 

A tweet is considered as relevant if it talks about anything that has to do with, but not limited to, the following issue categories: the Second Amendment, Gun control laws, etc. Tweets which contain the following hashtags are probably relevant to gun control: \#NoBillNoBreak, \#WearOrange, \#EndGunViolence, \#DisarmHate, \#molonlabe, etc.

A tweet should be considered as irrelevant if it mentions a gun death event or a gun violence news, but the context is not necessarily about gun control.

Some examples for relevant tweets and their stance (we only show the text here, but you need to answer this question from both the text and image): 
\begin{itemize}
[leftmargin=*]
\setlength\itemsep{0.00em}
    \item \textit{“Standing up for the second amendment and carrying a firearm for self defense.”} This tweet asks the audience to stand up for the 2nd amendment, which opposes to gun control;
    \item \textit{“I don’t understand why we can’t ban assault weapons. We all know they are only used for hunting people. \#PrayForOrlando \#guncontrolplease.”} This tweet talks about banning weapons and contains the hashtag “\#guncontrolplease”, which supports gun control;
    \item \textit{A common way to reduce violence in schools is to implement stronger security measures, such as surveillance cameras, security systems, campus guards and metal detectors. \#violence \#domesticviolence \#gun \#gunviolence \#abuse \#people \#world \#person \#workplace.”} This tweet is relevant to the topic, but we are not sure about its stance.
\end{itemize}

Some examples for non-relevant tweets (we only show the text here, but you need to answer this question from both the text and image): 
\begin{itemize}
[leftmargin=*]
\setlength\itemsep{0.00em}
    \item \textit{“Love will always conquer hate. \#PrayForOrlando \#OrlandoShooting.”} This tweet talks about gun violence but not about gun control; 
    \item \textit{“\#Gunviolence has serious and lasting social and emotional impacts on those who directly and indirectly experience it.”} This tweet points out the impact of gun violence but not about gun control.
\end{itemize}

\subsubsection{Stance: Immigration}
\begin{figure}[htbp]
    \centering
    \includegraphics[width=0.7\columnwidth]{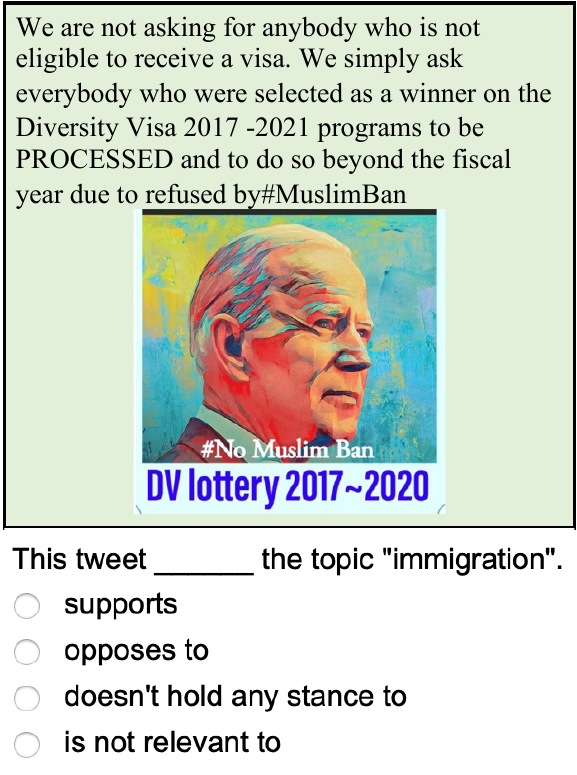}
    \caption{Example of stance annotation on immigration}
    \label{fig:stance_immigration}
\end{figure}

We aim to study the topic and the stance of tweets. Given a tweet accompanying with an image, you need to answer the stance of the tweet towards a given topic, as depicted in Figure \ref{fig:stance_immigration}.
Please make sure that you have the basic knowledge about that social topic and you understand the key message that the tweet (i.e. both the text and the image) sends. Just skip the HIT if you are not sure.

The question is about the stance. You need to decide whether the tweet is relevant or not to the social topic immigration. If it is relevant, then you need to annotate the stance: supports/opposes to/doesn’t hold any stance. 

A tweet is considered as relevant if it talks about anything that has to do with, but not limited to, the following issue categories: Borders, Birthright citizenship, Immigrant Crime, DACA and the DREAM Act, Deportation debate, Economic impact, Immigration quotas, Immigrants’ rights and access to services, Labor Market - American workers and employers, Law enforcement, Refugees, etc.

A tweet should be considered as irrelevant if it mentions a group of immigrant people such as Muslim, Syrian refugees but doesn’t explicitly talk about immigration issues. 

Some examples for relevant tweets and their stance (we only show the text here, but you need to answer this question from both the text and image): 
\begin{itemize}
[leftmargin=*]
\setlength\itemsep{0.00em}
    \item \textit{“Man feels bad for new immigrant driver in Brampton that crashed into his truck, causing \$6K worth of damages - he had no licence or insurance”.} This tweet is related to the topic of immigration under the category of Immigrant Crime, and it opposes to immigration.
    \item \textit{“House Bill 3438 will finally give our immigrant students some desperately needed resources! Thank you State Representative Maura Hirschauer for introducing this bill! Now, let's make sure this bill becomes law!”} This tweet is related to the topic of immigration under the category of DREAM Act, and it supports immigration.
    \item \textit{“I’m a woman that supports Trump to fix economy, immigration, school, military more. \#MAGA3X”} We consider a tweet as relevant even if it mentions several topics in addition to immigration, and it opposes to immigration.
\end{itemize}

Some examples for non-relevant tweets (we only show the text here, but you need to answer this question from both the text and image):
\begin{itemize}
[leftmargin=*]
\setlength\itemsep{0.00em}
    \item \textit{“’Will I die, miss?’ Terrified Syrian boy suffers suspected gas attack.”} This tweet talks about a Syrian boy suffering a gas attack, which may be pointing to a war or terrorist event in Syria, not necessarily directly about an immigration issue.
    \item \textit{“Virtual tour of Steinbach, in partnership with MANSO, Welcome Place, Eastman Immigrant Services and the Steinbach LIP, coming up March 9th, 2021. It's free so don't miss out!”} This tweet mentions Immigrant Services, but does not talk about any immigration issue.
    \item \textit{“I called on \@[USERNAME] for increased vaccine access for South Philadelphia seniors and for members of our immigrant communities. We can’t let physical distance and language barriers keep people from this lifesaving vaccine.”} This tweet talks about vaccine access for the immigrant community but it doesn’t hold any stance towards any immigration policy.
\end{itemize}

\subsubsection{Stance: Abortion}
\begin{figure}[htbp]
    \centering
    \includegraphics[width=0.7\columnwidth]{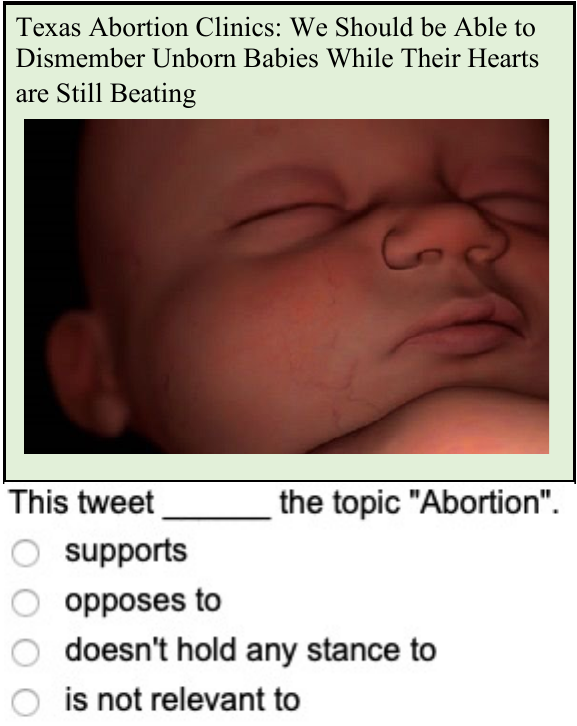}
    \caption{Example of stance annotation on abortion}
    \label{fig:stance_abortion}
\end{figure}

We aim to study the topic and the stance of tweets. Given a tweet accompanying with an image, you need to answer the stance of the tweet towards a given topic, as depicted in Figure \ref{fig:stance_abortion}.
Please make sure that you have the basic knowledge about that social topic and you understand the key message that the tweet (i.e. both the text and the image) sends. Just skip the HIT if you are not sure.

The question is about the stance. You need to decide whether the tweet is relevant or not to the social topic abortion. If it is relevant, then you need to annotate the stance: supports/opposes to/doesn’t hold any stance. 

A tweet is considered as relevant if it talks about anything that discusses whether the abortion should be a legal option. If the arguments in the tweet text and image support that the abortion should be a legal option, then please choose “supports”; if arguments oppose to legal abortion, then choose “opposes to”; if arguments doesn’t hold any stance for the topic then choose “doesn’t hold any stance”.
Notice that a tweet is considered as irrelevant if it doesn’t directly discuss whether the abortion should be a legal option or not, even though it may talk about related topics such as babies born alive after an abortion, birth control, etc.

\subsection{Persuasiveness level and image content} \label{Sec: content code manual}
\begin{figure}[htbp]
    \centering
    \includegraphics[width=0.7\columnwidth]{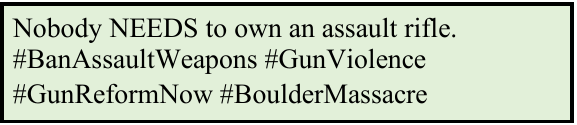}
    \caption{Example of a text only tweet}
    \label{fig:text_only}
\end{figure}

We aim to study the persuasiveness level of images in tweets as well as their content. Given a tweet text shown as Figure \ref{fig:text_only}, you need to give a persuasiveness score of it. 

\begin{figure}[htbp]
    \centering
    \includegraphics[width=0.7\columnwidth]{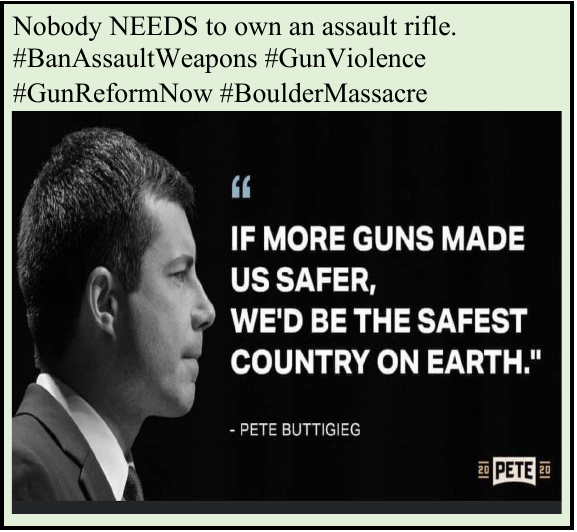}
    \caption{Example of a tweet accompanying an image}
    \label{fig:image}
\end{figure}

Then given a tweet accompanying an image shown as Figure \ref{fig:image}, you need to give a persuasiveness score again. 

Finally, you need to select the content type of the image. The content type of an image represents what type of the information the image mainly carries. Specifically, you need to pick one out of six types below for each image. 

\textbf{Statistics:} the image provides evidence by \textbf{stating or quoting quantitative information}, such as a chart/data analysis, that is related to the tweet text. 

An image could be considered statistics if: 1) It carries quantitative information (number/statistics/etc). 2) The key purpose of the image is to deliver this quantitative information, in the case there are multiple content types involved.

\begin{figure}[htbp]
    \centering
    \includegraphics[width=1\columnwidth]{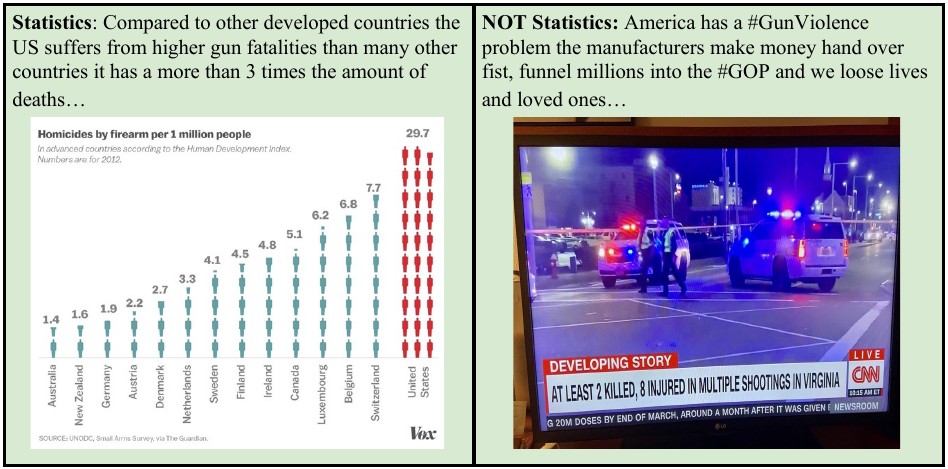}
    \caption{Example of tweets with statistics image and a non-statistics image.}
    \label{fig:statistics}
\end{figure}

For the examples shown in Figure \ref{fig:statistics}, in the statistics example, the image mainly shows a chart and delivers quantitative information (homicides by firearm per 1 million people). In contrast, in the NOT statistics example, though there are numbers in the image, the main information is a news title and the shooting scene, but not these numbers.

\textbf{Testimony:} the image \textbf{quotes statements or conclusions from an authority}, such as a piece of an article/claim from an official document, that is related to the tweet text. 

The image can be considered as testimony if: 1) The content contains texts such as statements/conclusions/pieces of article. 2) These texts are original from other resources such as news/celebrities/official documents/etc. 3) The key purpose of the image is to quote the authorized statement, in the case there are multiple content types involved.

\begin{figure}[htbp]
    \centering
    \includegraphics[width=1\columnwidth]{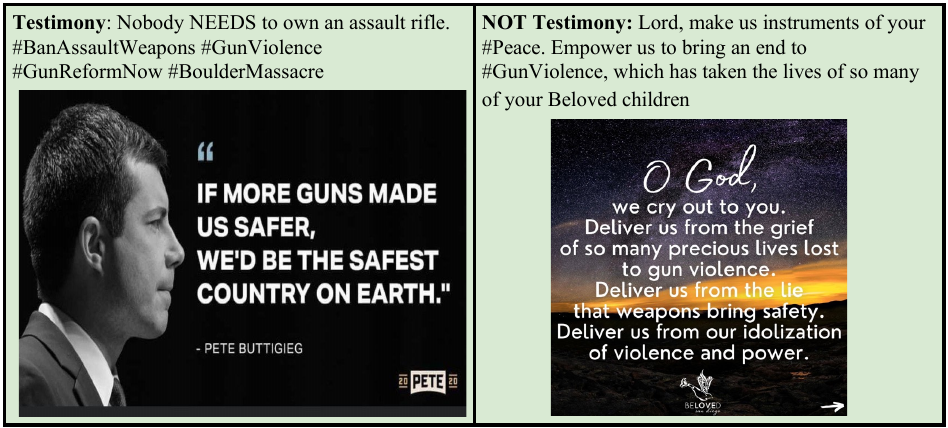}
    \caption{Example of tweets with testimony image and a non-testimony image.}
    \label{fig:testimony}
\end{figure}

For the examples shown in Figure \ref{fig:testimony}, in the Testimony tweet example, the image mainly cites a statement given by the transportation secretary. However, in the NOT Testimony tweet example, though it contains a piece of texts, these texts are not cited from an authority, therefore, it is not testimony.

\textbf{Anecdote:} the image provides information based on the \textbf{author's personal experience}, such as facts/personal stories, that are related to the tweet text. 

An image can be considered as an anecdote if: 1) It delivers a personal experience, Or 2) it shows a fact/experience that comes from personal view/known by the author. 3) The key purpose of the image is to deliver personal experience, in the case there are multiple content types involved.

\begin{figure}[htbp]
    \centering
    \includegraphics[width=1\columnwidth]{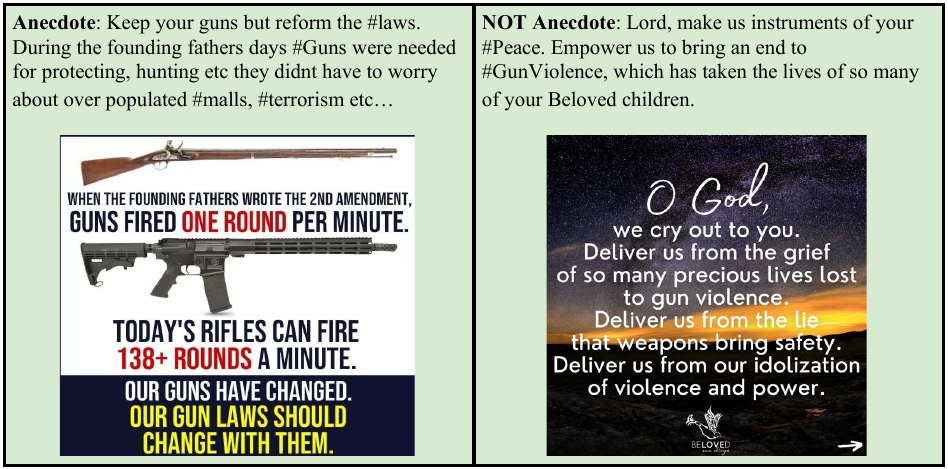}
    \caption{Example of tweets with anecdote image and a non-anecdote image.}
    \label{fig:anecdote}
\end{figure}

For the examples shown in Figure \ref{fig:anecdote}, the anecdote image shows the personal view on the fact that guns have been developed since the period of the 2nd amendment, and therefore the laws for guns should be developed as well. However, in the NOT anecdote example, though it comes from a personal statement, it does not describe any fact/experience/stories.

\textbf{Slogan:} the image expresses a piece of \textbf{advertising phrase}. 

An image can be considered as a slogan if: 1) It mainly delivers a piece of text as slogan; 2) The text is for advertising purposes as an advertising phrase/claim/statement. 3) The key purpose of the image is to deliver the piece of text, in the case there are multiple content types involved.

\begin{figure}[htbp]
    \centering
    \includegraphics[width=1\columnwidth]{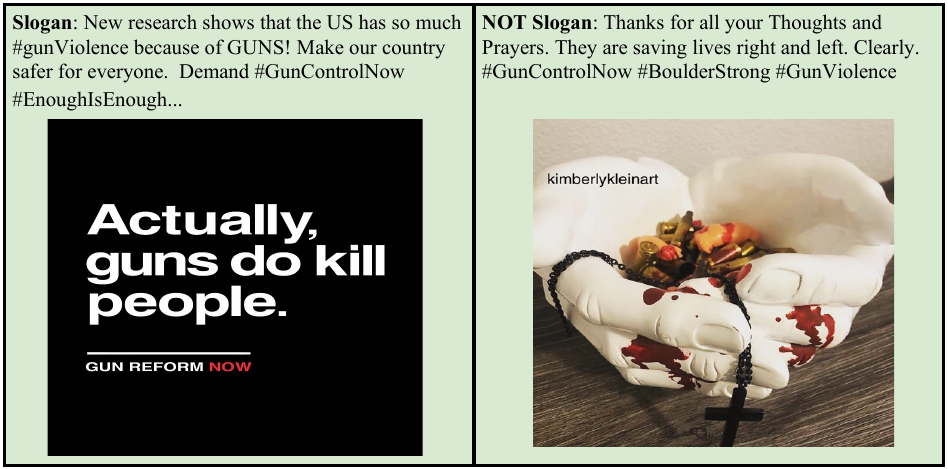}
    \caption{Example of tweets with slogan image and a non-slogan image.}
    \label{fig:slogan}
\end{figure}

For the examples shown in Figure \ref{fig:slogan}, the slogan image presents a phrase “Actually guns do kill people. Gun Reform Now”, therefore it is a slogan. However, For the 
example of NOT Slogan, though the image is for advertising, it does not contain a phrase for that, therefore it is not a slogan.

\textbf{Scene photo:} the image shows a \textbf{literal scene/photograph} that is related to the tweet text.

An image can be considered as a scene photo if: 1) It shows a literal photograph/scene. 2) The image is directly related to the text. 3) The key purpose of the image is to deliver the image content but not the text within, in the case there are multiple content types involved.

\textbf{Symbolic photo:} the image shows a \textbf{symbol/art} that expresses the author's viewpoints in a \textbf{non-literal} way. 

An image can be considered as a symbolic photo if: 1) It shows a symbol/art. 2) It expresses the viewpoint from the author in an implicit way. 3) The key purpose of the image is to deliver the image content but not the text within, in the case there are multiple content types involved.

\begin{figure}[htbp]
    \centering
    \includegraphics[width=1\columnwidth]{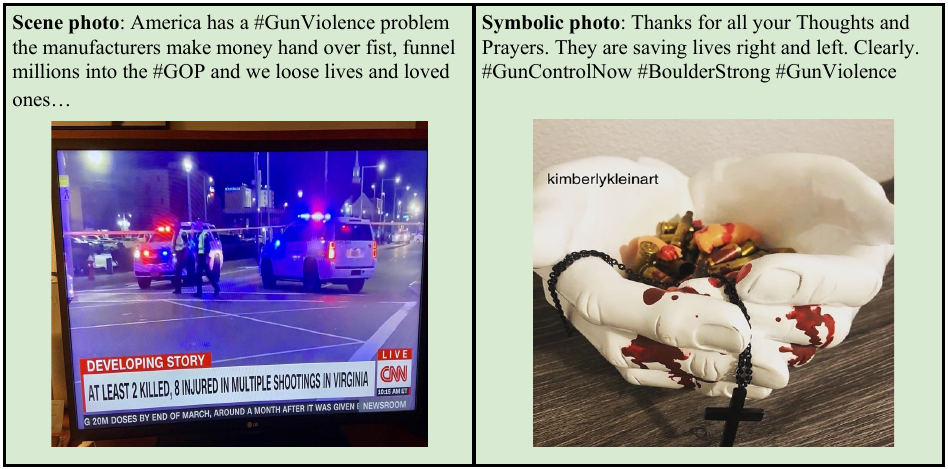}
    \caption{Example of tweets with scene photo image and a symbolic photo image.}
    \label{fig:scenephoto}
\end{figure}

For example, in Figure \ref{fig:scenephoto}, the scene photo image shows a real photograph of a gun violence scene reported by CNN news. In the Symbolic photo, though relevant to the text, it shows a photo/image that is related to the text in a non-literal way (blood signifies gun-killing and the hand posture signifies praying), therefore it is not a scene photo but a symbolic photo.

\begin{figure}[htbp]
    \centering
    \includegraphics[width=1\columnwidth]{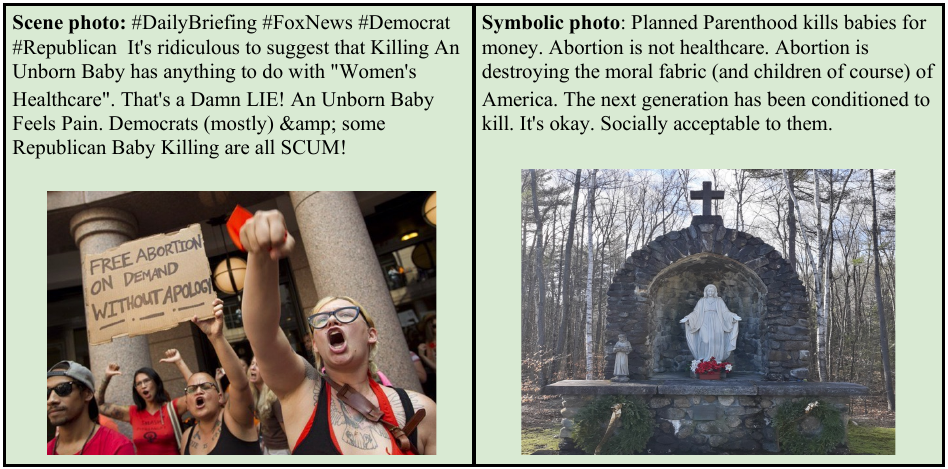}
    \caption{Another example of tweets with scene photo image and a symbolic photo image.}
    \label{fig:symbolicphoto}
\end{figure}

The key difference between the Scene photo and Symbolic photo is \textbf{whether the photograph sends a message literally or symbolically}. For a scene photo, the image directly expresses/supports the author’s view without any rhetoric; for a symbolic photo, the image may have several possible interpretations and the audience can understand its symbolic meaning after considering the tweet text. 
Consider the example shown in  Figure \ref{fig:symbolicphoto}: for the scene photo, it directly shows a protest scene and the author opposes to the abortion by considering it as a lie. In the symbolic photo, the author shows a photo of Notre Dame as a symbol of anti-abortion. The photo is not directly related to abortion, but audience can understand its symbolic meaning after reading the text.

\textbf{In the case there are multiple content types involved: }
You need to first identify the key purpose of the image (i.e. what is the most important information in the image). Then please select the content type of the key purpose. Table \ref{tab:content} shows the summary of content types for each key purpose.

\begin{table}[htbp]
\caption{Summary of content types for each key purpose}
\label{tab:content}
\begin{adjustbox}{width=1\columnwidth}
\begin{tabular}{|ll|l|}
\hline
\multicolumn{2}{|l|}{\textbf{Key Purpose}}                           & \textbf{Content Type}   \\ \hline
\multicolumn{2}{|l|}{Quantitative information in the image} & Statistics     \\ \hline
\multicolumn{1}{|l|}{\multirow{3}{*}{Textual information in the image}}   & Statements or conclusions from an authority & Testimony   \\ \cline{2-3} 
\multicolumn{1}{|l|}{}  & Personal experiences/views        & Anecdote       \\ \cline{2-3} 
\multicolumn{1}{|l|}{}  & Advertising phrases               & Slogan         \\ \hline
\multicolumn{1}{|l|}{\multirow{2}{*}{Graphical information in the image}} & Literal photograph                          & Scene Photo \\ \cline{2-3} 
\multicolumn{1}{|l|}{}  & Non-literal/rhetorical photograph & Symbolic Photo \\ \hline
\end{tabular}
\end{adjustbox}
\end{table}

\subsection{Persuasion Mode} \label{Sec: mode code manual}
We aim to study the \textbf{argumentative roles of images} in tweets. Given a tweet accompanying an image, we would ask you to choose the persuasion mode of the image. The persuasion mode of an image represents how the image convinces the audience. Specifically, we will ask you whether the image appeals to logic/emotion/credibility.  Additionally, we will ask you why you make the choices. 

\textbf{Q1:} Does the image make the tweet more persuasive by appealing to \textbf{logic and reasoning}?  

The image appeals to logic and reasoning if it persuades audiences with reasoning from a fact/statistics/study case/scientific evidence. Specifically, if: 1) the image \textbf{contains information for logic and reasoning}; 2) the image \textbf{presents logic and reasoning}.

Also, we will ask you why you made the choice. i.e. Describing the logic/reasoning brought by the image. Such as following, by filling the blank in the textbox:

\textit{The logic/reasoning of the image is  [the correlation between gun deaths and gun ownership by population].}

\begin{figure}[htbp]
    \centering
    \includegraphics[width=1\columnwidth]{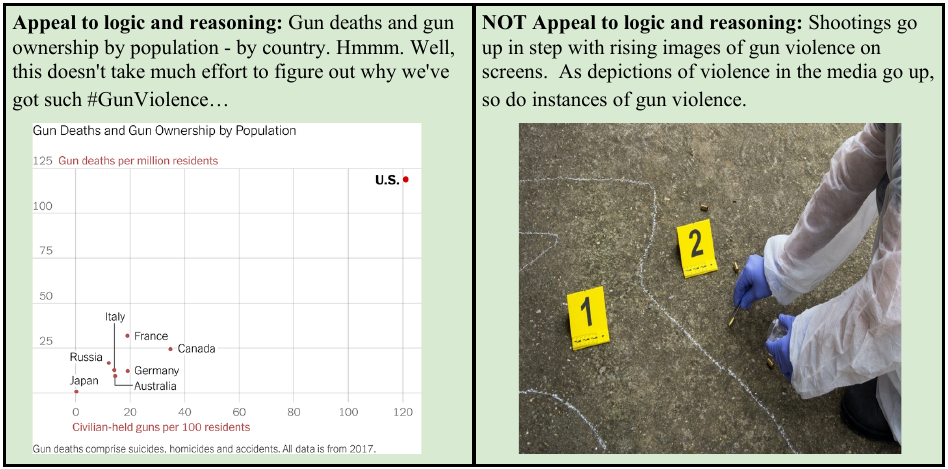}
    \caption{Example of tweets with logos image and non-logos image.}
    \label{fig:logos}
\end{figure}

For example shown in Figure \ref{fig:logos}, the left image provides a chart that shows the high gun deaths and the high gun ownership by the population of the US, which implies [a correlation between gun death and gun ownership which demonstrates that there will be less gun deaths with gun control.]. On the contrary, the right image shows the scene of the shooting but does not provide any reasoning or logic. 

\textbf{Q2:} Does image make the tweet more persuasive by appealing to \textbf{emotion}?  

The image appeals to \textbf{emotion}, if it puts audiences in a certain frame of mind by stimulating them to identify/empathize/sympathize with the arguments.

Specifically, if : 1) the image \textbf{invokes the audience with strong emotion}, such as sadness, happiness, compassion, worriness; 2) the image \textbf{makes the audience identify/empathize/sympathize} with the author/arguments.

Also, we will ask you why you made the choice. i.e. Describing the emotion(such as anger/amusement/sad/etc.) or impulsion(desire to do something) brought by the image. Such as following, by filling the blank within the [bracket]:

\textit{The image evokes my emotion/impulse of [anger].}

\begin{figure}[htbp]
    \centering
    \includegraphics[width=1\columnwidth]{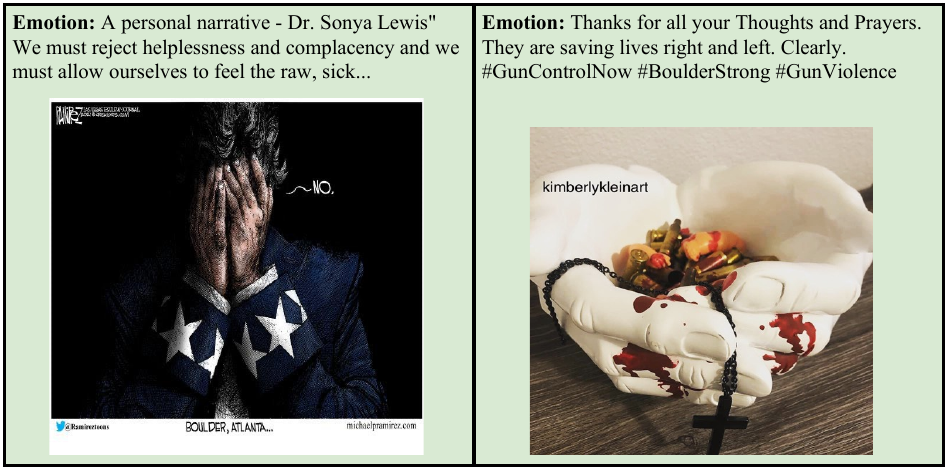}
    \caption{Example of tweets with pathos images.}
    \label{fig:pathos}
\end{figure}

For example shown in Figure \ref{fig:pathos}, the left image shows the grieved "Uncle Sam" saying "no" with helplessness, which evokes the [desire for gun control]. The right image provides an item that can revoke [compassion and forgiveness]. 

\textbf{Q3:} Does image make the tweet more persuasive by \textbf{enhancing credibility and trustworthiness}?  

The image \textbf{enhances credibility and trustworthiness}, if it makes people trust something more via authorized/trusted expertise/title/reputation. 

Specifically, if 1) The image \textbf{cites reliable sources} of the event/story/opinion/stance, that can make the contents trustworthy. Reliable sources include news, research reports, celebrated dictum, etc. Sources which are not proved/well-known by the audience (.e.g. an organization logo) are not considered as reliable. 2) the image \textbf{shows authorities} that can convince the audience to believe the arguments. 

Also, we will ask you why you made the choice. i.e. Describing the resources of the citation that enhances the credibility. Such as following, by filling the blank within the [bracket]:

\textit{The credibility is enhanced by [a citation to political report]}

\begin{figure}[htbpt]
    \centering
    \includegraphics[width=1\columnwidth]{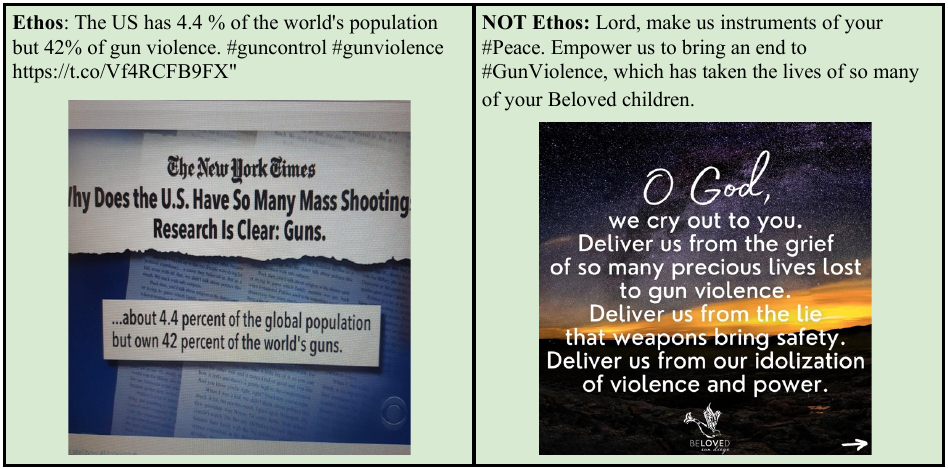}
    \caption{Example of tweets with ethos image and non-ethos image.}
    \label{fig:ethos}
\end{figure}

For example shown in Figure \ref{fig:ethos}, the left image takes a screenshot of the source of a report from [New York Times], which increases credibility. The NOT Ethos right image shows the views but are not quoted sentences that do not provide the credibility to enhance the argument.

\end{document}